\pdfoutput=1

\documentclass[11pt]{article}

\usepackage{EMNLP2023}

\usepackage{times}
\usepackage{latexsym}

\usepackage[T1]{fontenc}

\usepackage[utf8]{inputenc}

\usepackage{microtype}

\usepackage{inconsolata}

\usepackage{amsmath}
\usepackage{amssymb}
\usepackage{mathtools}
\usepackage{tabularx}
\usepackage{booktabs}
\usepackage{multirow}
\usepackage{bbding}
\usepackage{pifont}
\usepackage{wasysym}
\usepackage{amsfonts}
\usepackage{float}
\usepackage{mathtools, nccmath}
\usepackage{subcaption}

\usepackage{microtype}
\usepackage{graphicx}
\usepackage{caption}
\usepackage{subcaption}
\usepackage[inline,shortlabels]{enumitem}
\usepackage[ruled,vlined]{algorithm2e}
\usepackage{xurl} 

\graphicspath{{./figures}}

\newcommand{\en}{\textsc{en}}
\newcommand{\dutch}{\textsc{nl}}
\newcommand{\de}{\textsc{de}}
\newcommand{\es}{\textsc{es}}
\newcommand{\pt}{\textsc{pt}}
\newcommand{\italian}{\textsc{it}}
\newcommand{\ar}{\textsc{ar}}
\newcommand{\fr}{\textsc{fr}}
\newcommand{\zh}{\textsc{zh}}
\newcommand{\ja}{\textsc{ja}}
\newcommand{\ru}{\textsc{ru}}
\newcommand{\tu}{\textsc{tu}}
\newcommand{\hi}{\textsc{hi}}
\usepackage{geometry}
\usepackage[permil]{overpic}

\usepackage{tikz}
\usetikzlibrary{shapes}

\usepackage{siunitx}  
\usepackage[nameinlink]{cleveref}
\crefname{figure}{Fig.}{Figs.}
\Crefname{figure}{Figure}{Figures}
\crefname{equation}{Eq.}{Eqs.}
\Crefname{equation}{Equation}{Equations}
\crefformat{section}{#2§#1#3}
\crefname{section}{§}{§§}
\Crefname{section}{Section}{Sections}
\crefname{table}{Table}{Tables}
\crefname{appendix}{Appendix}{Appendices}

\newcommand{\etc}{etc.\ }
\newcommand{\eg}{e.g., }

\newcommand{\ie}{i.e., }
\newcommand{\vs}{vs.\ }
\newcommand{\etal}{\textrm{et al.\ }}

\usepackage{amsmath}
\usepackage{amssymb}
\usepackage{mathtools}
\usepackage{tabularx}
\usepackage{booktabs}
\usepackage{multirow}
\usepackage{bbding}
\usepackage{pifont}
\usepackage{wasysym}
\usepackage{amsfonts}
\usepackage{float}
\usepackage{mathtools, nccmath}
\usepackage{subcaption}

\usepackage{microtype}
\usepackage{graphicx}
\usepackage{caption}
\usepackage{subcaption}
\usepackage[ruled,vlined]{algorithm2e}
\usepackage{xurl} 

\graphicspath{{./figures}}

\usepackage{geometry}
\usepackage[permil]{overpic}

\usepackage{tikz}
\usetikzlibrary{shapes}

\usepackage{siunitx}  
\usepackage{colortbl}
\usepackage{xcolor, soul}
\definecolor{myorange}{RGB}{251, 230, 207}
\definecolor{myblue}{RGB}{220, 232, 250}
\definecolor{mybluedark}{RGB}{79, 92, 156}
\definecolor{myreddark}{RGB}{192, 84, 95}
\definecolor{mypink}{RGB}{241, 207, 205}
\definecolor{mygreen}{RGB}{216, 231, 214}
\definecolor{mygreendark}{RGB}{95, 157, 104}

\usepackage{tcolorbox}
\usepackage{pifont}
\usepackage{colortbl}

\newcommand{\win}{\textcolor{mygreendark}{$\uparrow$}}
\newcommand{\lose}{\textcolor{myreddark}{$\downarrow$}}

\usepackage{rotating}

\title{Zero-Shot Cross-Lingual Sentiment Classification under Distribution Shift: an Exploratory Study}

\author{{\centering 
    Maarten De Raedt$^{\diamondsuit\clubsuit}$~ 
    Semere Kiros Bitew$^\clubsuit$~
    Fréderic Godin$^\diamondsuit$~
    Thomas Demeester$^\clubsuit$~
    Chris Develder$^\clubsuit$}  \\
    $^\diamondsuit$ Sinch Chatlayer ~~$^\clubsuit$ Ghent University - imec \\
    \texttt{\{maarten.deraedt, semerekiros.bitew, thomas.demeester, chris.develder\}@ugent.be}  \\
    \texttt{frederic.godin@sinch.com} \\
}

\begin{document}
\maketitle
\begin{abstract}
The brittleness of finetuned language model
performance on out-of-distribution (OOD) test samples in unseen domains has been well-studied for English, yet
is unexplored for multilingual models.
Therefore, we study generalization to OOD test data specifically in zero-shot cross-lingual transfer settings, analyzing performance impacts of both \emph{language} and \emph{domain} shifts between train and test data.
We further assess the effectiveness of counterfactually augmented data (CAD) in improving OOD generalization for the cross-lingual setting, since CAD has been shown to benefit in a monolingual English setting.
Finally, we propose two new approaches for OOD generalization that avoid the costly annotation process associated with CAD, by exploiting the power of recent large language models (LLMs).
We experiment with 3 multilingual models, LaBSE, mBERT, and XLM-R trained on English IMDb movie reviews, and evaluate
on OOD test sets in 13 languages: Amazon product reviews, Tweets, and Restaurant reviews.
Results echo the OOD performance decline observed in the monolingual English setting.
Further,
\begin{enumerate*}[(i)]
\item counterfactuals from the original high-resource language do improve OOD generalization in the low-resource language, and
\item our newly proposed cost-effective approaches reach similar or up to to $+$3.1\% better accuracy than CAD
for Amazon and Restaurant reviews.
\end{enumerate*}
\end{abstract}

\section{Introduction}

To solve Natural Language Processing (NLP) tasks in low-resource languages, using multilingual models is a much adopted strategy \cite{devlin-etal-2019-bert, artetxe-schwenk-2019-massively, conneau2019cross, feng-etal-2022-language}.
A particularly popular paradigm is zero-shot cross-lingual transfer \cite{ruder2019survey, artetxe-etal-2020-cross, hu2020xtreme, lauscher-etal-2020-zero}: pre-trained multilingual models are finetuned on downstream tasks with training data solely from a high-resource language (\eg English). 
The resulting finetuned model can then be applied on a low-resource language samples, \ie without requiring costly training data in the low-resource language.

\begin{figure}[t]
    \centering
    \includegraphics[width=\columnwidth]{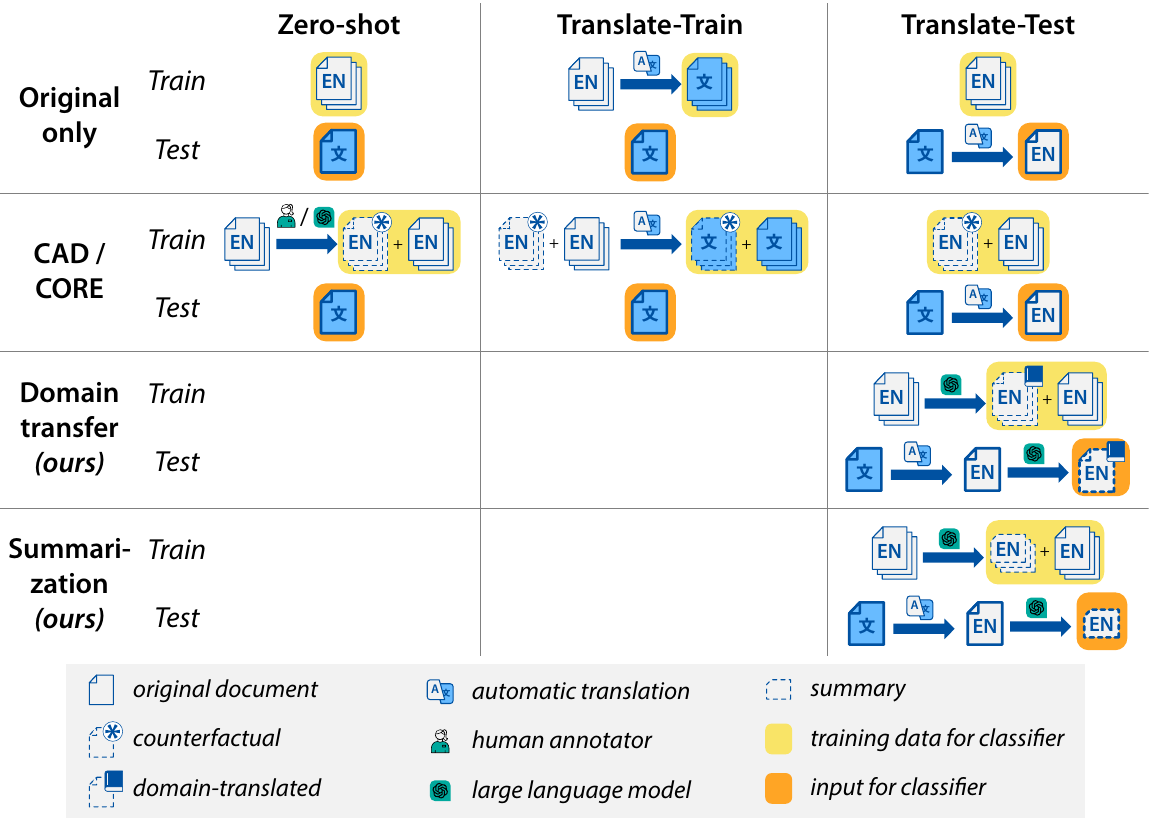}
    \caption{\textbf{Zero-shot cross-lingual transfer setup.} 
    Multiple transfer strategies, including 
    our newly proposed \emph{summarization} and \emph{domain transfer} methods for boosting OOD generalization.}
    \label{fig:illustration}
\end{figure}

In such zero-shot cross-lingual transfer, linguistic discrepancy between training and test languages causes a challenge:
typically, performance is sub-par compared to monolingual models.\footnote{Admittedly, such monolingual models do need low-resource training data.}
Several works have looked into narrowing the performance gap stemming from such language-based distribution shift \cite{liu-etal-2021-preserving, yu-joty-2021-effective,zheng-etal-2021-consistency, artetxe2023revisiting}.

Yet, besides the language-based shift, in real-world settings there may also be a domain-shift between training and test samples, \ie test samples may comprise out-of-distribution (OOD) data \cite{quinonero2008dataset}.
For example, a sentiment classifier to predict positive/negative appreciation by a consumer may be trained on movie reviews but applied on product reviews or tweets, where underlying sentiment features are assumed to be invariant \cite{arora-etal-2021-types}.

In a monolingual (English) setting, several studies unsurprisingly found a performance degradation when evaluating on OOD test data rather than on in-distribution (ID) data \cite{kaushik2019learning, kaushik2020explaining, gardner-etal-2020-evaluating, katakkar-etal-2022-practical}.
One of the underlying causes for that performance drop was found to be the classifier's reliance on spurious features, \ie patterns that from a human perspective should not be indicative for the classifier’s label \cite{poliak-etal-2018-hypothesis,gururangan-etal-2018-annotation,mccoy-etal-2019-right, wang-culotta-2020-identifying, joshi-etal-2022-spurious}: 
\eg \citet{wang-culotta-2020-identifying} found the 
occurrence of ``\emph{Spielberg}'' to be important for a positive sentiment classification.

A strategy that has been shown to improve OOD generalization in the monolingual English setting is the use of counterfactually augmented data (CAD), where annotators minimally revise training data to flip their labels \cite{kaushik2019learning}. 
Yet, constructing such annotations is costly: \citet{kaushik2019learning} report 5\,min/sample.

In this paper, we present an exploratory study of OOD generalization specifically in a \emph{cross-lingual} context, since we found this not to be covered in related work (\cref{sec:related}). 
Specifically, we 
\begin{enumerate*}[(i),afterlabel=\hskip 0.5pt]
    \item \label{it:xlood} identify the impact of OOD data on zero-shot \emph{cross-lingual} transfer performance, aiming to disentangle performance drops stemming from \emph{language} \vs \emph{domain} shifts between training and test data, and
    \item \label{it:xlcad} propose and analyze two new data augmentation strategies to improve OOD generalization that \emph{avoid the costly annotations} associated with using counterfactuals.
\end{enumerate*}
For both, we present an empirical study (\cref{sec:experiments}) within the domain of binary sentiment analysis.
We consider English IMDb reviews \cite{maas-etal-2011-learning} as in-distribution training data, with out-of-distribution test data spanning 13 languages across the Amazon \cite{keung-etal-2020-multilingual}, Tweets \cite{barbieri-etal-2022-xlm}, and Restaurants \cite{pontiki-etal-2016-semeval} datasets.
We further experiment with pre-trained multilingual models mBERT \cite{devlin-etal-2019-bert}, XLM-R \cite{conneau2019cross}, and LaBSE \cite{feng-etal-2022-language}.

For \ref{it:xlood}, we answer a first research question, \textbf{(RQ1)} \emph{How well do zero-shot cross-lingual methods trained with English sentiment data generalize to out-of-distribution samples in non-English languages?} 
To this end, we finetune the multilingual models on the English IMDb sentiment data, and evaluate their performance on OOD test samples in non-English languages.

For \ref{it:xlcad}, we answer \textbf{(RQ2)} \emph{How can zero-shot cross-lingual transfer methods better generalize to out-of-distribution samples, including for non-English languages?}
We will consider a CAD baseline as proposed by \citet{kaushik2019learning}, where annotators minimally revise training data to flip their labels, 
since training on both original and counterfactual data improves 
OOD generalization to unseen domains in the monolingual English setting. Specifically, we finetune the multilingual models on both the original English and counterfactually revised English IMDb reviews, and evaluate whether the OOD generalization gains observed in the monolingual setting translate also to OOD test samples in non-English languages. 

We then propose~(\cref{subsec:data_augmentation}) two cost-effective alternatives for CAD, using Large Language Models (LLMs):
\begin{enumerate*}[(1),afterlabel=\hskip 0.5pt]
    \item \label{it:domaintransfer} \emph{domain transfer}, and
    \item \label{it:summarization} \emph{summarization},
\end{enumerate*}
as illustrated in the 2 bottom rows of \cref{fig:illustration}.
For~\ref{it:domaintransfer}, we prompt an LLM to minimally edit both ID training and OOD test samples to map them onto the same, \emph{hypothetical} domain, \eg books.
For~\ref{it:summarization}, we prompt an LLM to abstractly summarize both ID training and OOD test data, since we hypothesize that summaries 
can capture the core essence of samples while removing non-essential, potentially spurious, information.

Our results~(\cref{sec:results}) show that in the 
OOD test setting for non-English languages, model performance of zero-shot cross-lingual transfer substantially declines, aligned with OOD generalization studies in a monolingual English setting.
We further find that CAD improves OOD generalization for non-English samples, with gains up to $+$14.8\%, $+$4.7\%, and $+$7.9\% for respectively LaBSE, mBERT, and XLM-R. 
Finally, our cost-effective \emph{domain transfer} and \emph{summarization} data augmentation methods similarly improve OOD generalization, on par with or even surpassing CAD for \emph{Amazon} and \emph{Restaurants} by up to $+$3.1\% in accuracy.

\section{Related Work}
\label{sec:related}

\paragraph{Zero-shot cross-lingual transfer:}
A large part of multilingual NLP research focuses on improving the transfer of multilingual models trained on high-resource language data 
to low-resource languages. This can be achieved either by 
\begin{enumerate*}[(i)]
    \item cross-lingual pre-training schemes that yield stronger multilingual models \cite{artetxe-schwenk-2019-massively, conneau2019cross, conneau-etal-2020-unsupervised,xue-etal-2021-mt5, feng-etal-2022-language, chi-etal-2022-xlm}, or
    \item finetuning strategies that facilitate better cross-lingual transfer \cite{liu-etal-2021-preserving, yu-joty-2021-effective,zheng-etal-2021-consistency}.
\end{enumerate*}
Recently,
\citet{artetxe2023revisiting} revisited the \emph{translate-test} and \emph{translate-train} baselines \cite{shi-etal-2010-cross, duh-etal-2011-machine, artetxe-etal-2020-translation}, where \emph{test} samples are translated into English prior to evaluating them, or, respectively, the \emph{training} samples are translated into the test languages for fine-tuning a multilingual model.
Artetxe \etal found that using more recent machine translation systems, \eg NLLB \cite{costa2022no}, further boosts performance and often surpasses strong zero-shot cross-lingual methods.
Hence, we also experiment 
with \emph{translate-test} and \emph{translate-train} approaches.

\paragraph{Cross-lingual transfer under distribution shift:}
The limited research on the robustness of multilingual models has primarily focused on being robust against specific types of \emph{noise},
\eg adversarial perturbations for Japanese Natural Language Inference \cite{yanaka-mineshima-2021-assessing}, a combination of general and task-specific text transformations based on manipulating synonyms, antonyms, syntax, \etc \cite{wang-etal-2021-textflint}, and introducing errors and noise through Wikipedia edits \cite{cooper-stickland-etal-2023-robustification}.
Unlike these works, we will evaluate how well zero-shot cross-lingual transfer from English to non-English test samples can generalize in scenarios where
there is a shift in \emph{domain} from train to test data: the domain-specific features of test samples may change, whereas the semantic sentiment features remain invariant.

\paragraph{Counterfactually-augmented data (CAD):}
For English sentiment analysis, CAD is widely adopted to mitigate the effect of spurious patterns.
For example, \citet{kaushik2019learning, kaushik2020explaining} recruited Mechanical Turk workers to construct counterfactually revised samples by flipping labels with minimal editing, helping classifiers to learn real associations between samples and labels, thereby improving OOD generalization to unseen test domains. 
Building upon the success of CAD, several works have also studied how to automatically generate counterfactuals for English sentiment analysis \cite{wang2021robustness, yang-etal-2021-exploring, dixit-etal-2022-core, howard-etal-2022-neurocounterfactuals, de-raedt-etal-2022-robustifying}.
We adopt this CAD idea for OOD generalization in a zero-shot cross-lingual setting, which to the best of our knowledge has not been studied yet.

We start by exploring whether augmenting the English training data with the manually constructed counterfactuals from \citet{kaushik2019learning} also benefits OOD generalization for non-English test samples.
Additionally, we propose two new LLM-based methods as alternatives to constructing counterfactuals, aiming to specifically improve zero-shot transfer to non-English OOD test samples.
We benchmark our new LLM-based methods against a CAD setup following \citet{kaushik2019learning}, thus assessing whether we can achieve similar OOD performance but avoid CAD's costly human annotations. 
We further contrast classifiers trained on data augmented by our two new LLM-based methods to those trained on counterfactuals generated by CORE \cite{dixit-etal-2022-core}, the state-of-the-art method in automatic counterfactual generation. CORE first retrieves naturally occurring counterfactual edits from an unlabeled text corpus and then, based on these retrieved edits, instructs an LLM (GPT-3) to counterfactually revise training samples.

\section{Experimental Setup}
\label{sec:experiments}
We describe the English ID training data and non-English OOD test data in \cref{subsec:datasets}. Next, we outline the pre-trained multilingual models and the transfer strategies we experiment with in \cref{subsec:transfer}. In \cref{subsec:data_augmentation}, we present our LLM-based \emph{domain transfer} and \emph{summarization} data augmentation methods. We cover finetuning and evaluation in \cref{subsec:finetuning_evaluation}.

\begin{table*}[t!]
\smaller
\scriptsize
\centering
\addtolength{\tabcolsep}{-3pt}
\begin{tabular}{cp{2cm}p{12.5cm}}
\toprule
& & \textbf{Original samples} \\ 
\multirow{4}{*}{\includegraphics[width=.5cm]{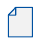}} & \textsc{imdb} & If you haven't \textcolor{blue!70}{seen} this, it's terrible. It is pure trash. I \textcolor{blue!70}{saw} this about 17 years ago, and I'm still screwed up from it.
\\
& \textsc{tweets} &
She just didn't get \textcolor{blue!70}{them} in areas were she needed them. Lots of \textcolor{blue!70}{voter supression} going on. \textcolor{blue!70}{Hacking} \& \textcolor{blue!70}{tampering}
\\
& \textsc{amazon} & The \textcolor{blue!70}{straps} are super \textcolor{blue!70}{small}, for a very \textcolor{blue!70}{small wrist}, and the \textcolor{blue!70}{closure} is \textcolor{blue!70}{bad}, easy to lose the \textcolor{blue!70}{watch}.
\\ 
& \textsc{restaurants} & The \textcolor{blue!70}{food} is \textcolor{blue!70}{standard}, but the person waiting at the \textcolor{blue!70}{door} in the style of a manager is cold and unfriendly.
\\
\midrule
& & \textbf{Domain transferred samples} \\
\multirow{4}{*}{\includegraphics[width=.5cm]{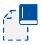}} &
\textsc{imdb} & If you haven't \textcolor{red!70}{read} this \textcolor{red!70}{book}, it's terrible. It is pure trash. I \textcolor{red!70}{read} this about 17 years ago, and I'm still screwed up from it.
\\ 
& \textsc{tweets} & She just didn't get the \textcolor{red!70}{books} in areas where she needed them. Lots of \textcolor{red!70}{book censorship} going on. \textcolor{red!70}{Piracy} \& \textcolor{red!70}{Plagiarism}
\\ 
& \textsc{amazon} & The \textcolor{red!70}{binding of the book} is super \textcolor{red!70}{tight}, suited for a \textcolor{red!70}{compact size}, and the \textcolor{red!70}{cover} is \textcolor{red!70}{not secure}, easy to lose the \textcolor{red!70}{pages}.
\\
& \textsc{restaurants} & The \textcolor{red!70}{books} are \textcolor{red!70}{average}, but the person at the \textcolor{red!70}{front desk} in a manager-like role is distant and unapproachable.
\\
\midrule
& & \textbf{Summarized samples}
\\
\multirow{4}{*}{\includegraphics[width=.5cm]{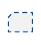}} &
\textsc{imdb} & Terrible and traumatizing movie, avoid it.
\\
& \textsc{tweets} & Allegations of voter suppression and tampering.
\\ 
& \textsc{amazon} & Small straps, bad closure, easy to lose.
\\
& \textsc{restaurants} & Standard food, unfriendly manager.
\\
\bottomrule
\end{tabular}
\addtolength{\tabcolsep}{3pt}
\caption{\textbf{LLM-based data-augmentation.} \emph{Top:} original ID training and OOD test samples (including English translations).  \emph{Middle:} mapping of the diverse domain samples onto the \emph{hypothetical} books domain. \emph{Bottom:} demonstrates how \emph{summarization} retains essential information while removing potentially spurious elements.}
\label{table:illustration}
\end{table*}

\subsection{Datasets}
\label{subsec:datasets}

\paragraph{In-distribution (ID) training data:}
We use the subset of 1,707 English reviews selected by \citet{kaushik2019learning} from the IMDb sentiment dataset \cite{maas-etal-2011-learning} as training data, as well as 245 English validation samples. To better assess the OOD generalization of cross-lingual transfer, we also report in-distribution results of all 13 considered languages on the IMDb test set with 488 samples. However, the test set of \citet{kaushik2019learning} is English-only.
Hence, we translate the 488 English test samples into each of the 12 other non-English languages, using OpenAI's \texttt{ChatGPT-turbo (v0301)} \cite{ouyangtraining}, 
as it achieves translation quality that is competitive to commercial machine translation tools (\eg Google Translate or Microsoft Translation Suite)  \cite{jiao2023chatgpt, hendy2023good, Peng2023ChatGPT4MT}, while being more cost-effective.
Since we aim to explore the benefits of English CAD for OOD generalization also to non-English test samples, we augment the respectively 1,707 and 488 original training and validation samples with their English counterfactually revised counterparts provided by \citet{kaushik2019learning}.
All training, validation, and test sets are equally balanced between positive and negative samples. 

\paragraph{Out-of-distribution (OOD) test data:}
Our OOD test data comprises three non-movie domains: \emph{product reviews}, 
\emph{tweets} and \emph{restaurant feedback}.
We use the MARC dataset \cite{keung-etal-2020-multilingual} for Amazon \emph{product reviews} in six languages: English, German, French, Spanish, Japanese, and Chinese.
For \emph{tweets}, we use the recent multilingual test sets provided by \citet{barbieri-etal-2022-xlm},
in eight languages: English, German, French, Spanish, Arabic, Hindi, Portuguese, and Italian. 
For \emph{restaurant} reviews, we use
the multilingual aspect-based sentiment classification dataset for the 2016 SemEval Task~5 \cite{pontiki-etal-2016-semeval}, \ie its restaurant domain data covering six languages: English, Dutch, French, Spanish, Russian, and Turkish.
Since SemEval Task~5 concerns aspect-based sentiment, we apply a rule-based mapping to cast it as a binary classification task: included reviews are labeled either as \emph{positive} (if all aspects are positive or a mix of neutral and positive) or \emph{negative} (if all aspects are negative or a mix of neutral and negative).
We undersample test examples from the majority sentiment to ensure that all test sets are balanced.
Further dataset statistics are provided in \cref{sec:appendix}.

\subsection{Zero-shot cross-lingual transfer}
\label{subsec:transfer}
\paragraph{Pre-trained multilingual models:} 
We consider the base-cased versions of two multilingual language models pre-trained on masked language model (MLM) objectives: mBERT, \ie a multilingual variant of BERT \cite{devlin-etal-2019-bert}, and XLM-R, a RoBERTa-based multilingual model \cite{conneau2019cross}. Additionally, we use the pre-trained multilingual sentence encoder LaBSE \cite{feng-etal-2022-language} that maps sentences to 768-dimensional single vector representations. 

\paragraph{Transfer strategies:}
To transfer from the English ID training data to non-English test samples, we use 3 widely adopted strategies (\cref{fig:illustration}, top row):
\begin{enumerate}[(1),wide,labelindent=0pt,nosep]
    \item \emph{zero-shot}: finetunes the multilingual model on the English ID training and validation set, followed by directly evaluating  the OOD test samples in the non-English languages. 
    \item \emph{translate-test}: finetunes the multilingual model on the English ID training and validation datasets. However, prior to making predictions for OOD test samples, the samples are
    translated into English. 
    \item \emph{translate-train}: first translates the English ID training and validation datasets to the target OOD test language. Subsequently, the multilingual model is trained on this translated data to then make predictions for the original, untranslated, OOD test samples in that non-English language.
\end{enumerate}

Note that in case where  both \emph{translate-train}  and CAD are used, the English CAD training and validation data are translated to the target OOD test language. For both \emph{translate-test} and \emph{translate-train}, we use  OpenAI's   \texttt{ChatGPT-turbo (v0301)} \cite{ouyangtraining} as the LLM to translate from English to non-English languages and vice versa. We adopt OpenAI's default parameter values. See \cref{sec:appendix} for translation prompts.

\subsection{LLM-based data-augmentation}
\label{subsec:data_augmentation}
We explore whether data augmentation using an LLM, as a cost-effective alternative to CAD, can also boost OOD generalization. We propose two such alternatives: 
\begin{enumerate*}[(1)]
    \item \emph{domain transfer}, and 
    \item \emph{summarization}.
\end{enumerate*} 
Our focus is on augmenting data for \emph{translate-test}, as recent work has shown it to be more effective than \emph{zero-shot} and \emph{translate-train} \cite{artetxe2023revisiting}. The multilingual models are finetuned on the original English ID, as well as the augmented ID training samples\footnote{To ensure all strategies have the same number of training samples, we train the \emph{original-only} models (without manual counterfactuals or LLM-augmented samples) on twice the number (3.4k) of original IMDb reviews (\cref{subsec:finetuning_evaluation}).}, with predictions made solely on augmented test samples. \Cref{table:illustration} provides illustrations for both strategies.

\paragraph{Domain transfer:} We align the domains of both the original ID training and OOD test samples \emph{translated} into English to a common \emph{hypothetical} domain.
To achieve this, we instruct \texttt{ChatGPT-turbo (v0301)} \cite{ouyangtraining} to minimally change the samples so that they relate to the new \emph{hypothetical} domain, for which we experiment with the domain of \emph{books}.
Note that rather than solely mapping OOD test samples to the ID training domain of movies, we use a \emph{hypothetical} domain to transform both training and test samples with an LLM to avoid introducing a new distribution shift caused by the mismatch between the original human-based training and the LLM-generated test samples.
See \cref{sec:appendix} for our domain transfer prompt.

\paragraph{Summarization:}
For our second augmentation strategy, we abstractly summarize both the original English training and the \emph{translated} English OOD test samples.
We hypothesize that such concise summaries can retain essential information while omitting non-essential and potentially spurious features, such as, \eg specific syntax structures and lexical choices, thereby a priori preventing classifiers from relying on such features for prediction.
Furthermore, transforming text with language models, \ie through summarization, may have the added benefit of normalizing the background, non-sentiment related, features.
Hence, summarizing the data can
lead to more uniform syntax and word choice among test and training samples, potentially further narrowing the distribution mismatch between ID training and OOD test samples.
\Cref{sec:appendix} lists the exact prompt that we supply to \texttt{ChatGPT-turbo (v0301)} \cite{ouyangtraining}, using OpenAI's default parameter values.

\subsection{Finetuning and evaluation}
\label{subsec:finetuning_evaluation}

We finetune the MLM-based models, \ie mBERT and XLM-R, by adding a classification head to the $\lbrack$\texttt{CLS}$\rbrack$-token. 
We use the Hugging Face Transformers library \cite{wolf-etal-2020-transformers} and train on a single Tesla V100 GPU for 20 epochs, with a batch size of 38, and a learning rate of \num{5e-6}.
To select an optimal model, we employ early validation stopping with a loss threshold of 0.01 and a patience of 10.
Since we are also interested in measuring the performance of a more compute-efficient model, we freeze LaBSE's parameters and train on CPU a logistic regression model on LaBSE's sentence vectors through five-fold cross-validation.
We use the \texttt{scikit-learn} library \cite{pedregosa2011scikit}, with \texttt{lbfgs} \cite{liu1989limited} as the solver, and set the maximum number of iterations to 5,000.

To assess the performance of each transfer strategy, we report the mean accuracy over 5 randomly initialized training runs, \ie with randomly selected weights and cross-validation folds for respectively mBERT/XLM-R and LaBSE.

Note that classifiers trained on CAD, as well as on data augmented by our two strategies, use respectively 1.7k extra manually constructed counterfactuals and 1.7k extra LLM-generated samples, in addition to the 1.7k original IMDb training samples. 
To ensure that the OOD generalization gains from CAD and our two augmentation strategies are not solely attributed to the increased number of training samples, we randomly sample an extra 1.7k original English IMDb reviews from the IMDb dataset of \citet{maas-etal-2011-learning} for the \emph{original-only} strategy (\ie models trained without counterfactuals or augmented data). As such, all considered strategies are trained on 3.4k samples

\section{Experimental Results and Discussion}
\label{sec:results}

\begin{table*}[h!]
\smaller
\centering
\begin{tabular}{l  cc cc cc cc}
\toprule
& \multicolumn{2}{>{\columncolor{blue!5}}c}{\textsc{imdb}} & \multicolumn{2}{>{\columncolor{red!5}}c}{\textsc{amazon}} & \multicolumn{2}{>{\columncolor{red!5}}c}{\textsc{restaurants}} & \multicolumn{2}{>{\columncolor{red!5}}c}{\textsc{tweets}}\\
\cmidrule(lr){2-3}  \cmidrule(lr){4-5} \cmidrule(lr){6-7} \cmidrule(lr){4-5} \cmidrule(lr){8-9} 
Method & \en & \textsc{non-en} & \en & \textsc{non-en} & \en & \textsc{non-en} &  \en & \textsc{non-en} \\ \midrule
\multicolumn{9}{l}{\textbf{LaBSE}} \\
- \textsc{zshot} & 85.0 & 84.9 & 66.3 & 71.9 & 72.7 & 74.1 & 76.3 & 67.8 \\
- \textsc{ttrain} & 85.0 & 85.2 & 66.3 & 74.0 & 72.7 & 76.4 & 76.3 & 66.0 \\
- \textsc{ttest} & 85.0 & - & 66.3 & 67.6 & 72.7 & 73.1 & 76.3 & 68.8 \\
\midrule
\multicolumn{9}{l}{\textbf{mBERT}} \\
- \textsc{zshot} & 89.5 & 80.8 & 79.3 & 72.2 & 80.2 & 69.6 & 75.9 & 62.8  \\
- \textsc{ttrain} & 89.5 & 87.5 & 79.3 & 73.5 & 80.2 & 74.5 & 75.9 & 62.9 \\
- \textsc{ttest} & 89.5 & - & 79.3 & 77.8 & 80.2 & 78.9 & 75.9 & 71.1 \\
\midrule
\multicolumn{9}{l}{\textbf{XLM-R}} \\
- \textsc{zshot} & 92.4 & 88.4 & 86.3 & 85.0 & 86.0 & 79.2 & 84.3 & 69.2  \\
- \textsc{ttrain} & 92.4 & 90.7 & 86.3 & 86.0 & 86.0 & 83.0 & 84.3 & 72.5 \\
- \textsc{ttest} & 92.4 & - & 86.3 & 85.6 & 86.0 & 81.5 & 84.3 & 71.7 \\
\bottomrule
\end{tabular}
\caption{
\textbf{\colorbox{blue!5}{In-distribution} \vs \colorbox{red!5}{out-of-distribution}} test accuracies for the \emph{original only} strategy trained solely on IMDb reviews (without CAD or data augmentation). Results are presented for English (\en) and non-English (\textsc{non-en}) test data, with the latter's accuracies averaged across all non-English languages per test set. Detailed results per language are provided in \cref{sec:appendix}. Note, for English, \textsc{ttrain} and \textsc{ttest} do not involve any translation, hence their \en\ scores are equivalent to \textsc{zshot}. Further, ID scores for \textsc{ttest} are omitted as these would involve backtranslating the non-English ID samples (originally translated from English ID test data per \cref{subsec:datasets}) to English, which would largely assess back-translation quality.}
\label{table:RQ1}
\end{table*}

\subsection{Zero-shot cross-lingual out-of-distribution generalization}\label{subsec:results_rq1}
We first address \textbf{(RQ1)}, on assessing OOD generalization to non-English samples. In \cref{table:RQ1}, we present both ID and OOD accuracies of the \textit{original only} method, which trains solely on (translated) English IMDb movie reviews without data augmentation.

We see that both for English and non-English, all models and transfer strategies decline in performance when evaluated on OOD  rather than ID test samples. For example, the \emph{zero-shot} strategy's drop from English ID to English OOD (ID\textsubscript{\textsc{en}}$\to$OOD\textsubscript{\textsc{en}}) ranges from 8.7\%--18.7\% for LaBSE, 9.3\%--13.6\% for mBERT, and 6.1\%--8.1\% for XLM-R. Similarly, for non-English (ID\textsubscript{\textsc{non-en}}$\to$OOD\textsubscript{\textsc{non-en}}), the performance drops for LaBSE, mBERT, and XLM-R vary within the ranges of 10.8\%--17.1\%, 8.6\%--18\%, and 3.4\%--19.2\%, respectively. These findings suggest that model performance decline to OOD test samples in non-English is substantial, as was already known (and here confirmed again) for English. We do not, however, see a consistently stronger decline for non-English than for English, as may be intuitively expected. This is discussed in more detail in the next paragraph.

\paragraph{English \vs non-English OOD generalization:}
 We assess whether multilingual models generalize better to English than non-English OOD test data.
 Overall, the \textsc{en} versus \textsc{non-en} scores in \cref{table:RQ1} reveal that the MLM-based models mBERT and XLM-R generalize less well to non-English compared to English OOD test samples: the accuracies for non-English languages are lower in most cases. Surprisingly, the converse holds for LaBSE: it has consistently better non-English OOD accuracies compared to English on \emph{Amazon} and \emph{Restaurants}. 
 Note, however, the absolute performance of the three models: LaBSE appears to be the least accurate model in English in most cases. This is consistent with the fact that its encoder remains frozen during training in English, unlike the other encoders, whereas LaBSE's non-English performance is more on par with the other models. While our results suggest that performance decline to OOD test samples in non-English and English is substantial, the disparity among OOD model performance between non-English and English depends on the
\begin{enumerate*}[(i)]
    \item pre-trained multilingual model or finetuning strategy, and
    \item the type of OOD data.
\end{enumerate*}

\begin{table*}[t!]
\scriptsize
\centering
\addtolength{\tabcolsep}{-4.05pt}
\begin{tabular}{l cccccc cccccc cccccc}
\toprule
& \multicolumn{6}{c}{\textbf{LaBSE}} & \multicolumn{6}{c}{\textbf{mBERT}} & \multicolumn{6}{c}{\textbf{XLM-R}} \\
\cmidrule(lr){2-7}  \cmidrule(lr){8-13} \cmidrule(lr){14-19} 
 & \multicolumn{2}{c}{\textsc{amazon}} & \multicolumn{2}{c}{\textsc{restaurants}} & \multicolumn{2}{c}{\textsc{tweets}} & \multicolumn{2}{c}{\textsc{amazon}} & \multicolumn{2}{c}{\textsc{restaurants}} & \multicolumn{2}{c}{\textsc{tweets}} & \multicolumn{2}{c}{\textsc{amazon}} & \multicolumn{2}{c}{\textsc{restaurants}} & \multicolumn{2}{c}{\textsc{tweets}} \\ \cmidrule(lr){2-3}  \cmidrule(lr){4-5} \cmidrule(lr){6-7} \cmidrule(lr){8-9} \cmidrule(lr){10-11} \cmidrule(lr){12-13} \cmidrule(lr){14-15} \cmidrule(lr){16-17}  \cmidrule(lr){18-19} 
Method & \en & \textsc{non-en} & \en & \textsc{non-en} & \en & \textsc{non-en} & \en & \textsc{non-en} & \en & \textsc{non-en} & \en & \textsc{non-en} & \en & \textsc{non-en} & \en & \textsc{non-en} & \en & \textsc{non-en} \\ \midrule
\multicolumn{11}{l}{\textbf{Original only}} \\
- \textsc{zshot} & 66.3 & 71.9 & 72.7 & 74.1 & 76.3 & 67.8 & 79.3 & 72.2 & 80.2 & 69.6 & 75.9 & 62.8 & 86.3 & 85.0 & 86.0 & 79.2 & 84.3 & 69.2  \\
- \textsc{ttrain} & 66.3 & 74.0 & 72.7 & 76.4 & 76.3 & 66.0 & 79.3 & 73.5 & 80.2 & 74.5 & 75.9 & 62.9 & 86.3 & 86.0 & 86.0 & 83.0 & 84.3 & 72.5 \\
- \textsc{ttest} & 66.3 & 67.6 & 72.7 & 73.1 & 76.3 & 68.8 & 79.3 & 77.8 & 80.2 & 78.9 & \underline{75.9} & 71.1 & 86.3 & 85.6 & 86.0 & 81.5 & \underline{84.3} & 71.7  \\
\midrule
\multicolumn{11}{l}{\textbf{Original + CAD} \cite{kaushik2019learning}} \\
- \textsc{zshot} & 81.2 & \underline{82.9} & 84.7 & 85.7 & 81.7 & \underline{74.5}  & 81.7 & 74.9 & 81.8 & 70.9 & 79.0 & 67.2 & 87.0 & 85.7 & 87.5 & 81.9 & 86.7 & 75.9  \\
- \textsc{ttrain} & 81.2  & 82.3 & 84.7 & 83.4 & 81.7 & 73.7  & 81.7 & 78.2 & 81.8 & 75.7 & 79.0 & 66.9 & 87.0 & 86.4 & 87.5 & 84.6 & 86.7 & 77.3 \\
- \textsc{ttest} & 81.2  & 82.4 & 84.7 & 85.9 & \textbf{81.7} & \textbf{76.2} & \textbf{81.7} & \textbf{81.2} & 81.8 & \underline{81.2} & \textbf{79.0} & \textbf{75.0} & 87.0 & 86.8 & 87.5 & 87.1 & \textbf{86.7} & \underline{79.6} \\
\midrule
\multicolumn{11}{l}{\textbf{Original + CORE} \cite{dixit-etal-2022-core}} \\
- \textsc{zshot} & 81.0 & 82.0 & 85.0 & 84.9 & 77.4 & 71.1  & 80.2 & 74.1 & 80.4 & 69.6 & 73.6 & 64.8 & 86.8 & 87.0 & 89.7 & 87.5 & 83.9 & 77.9 \\
- \textsc{ttest} & \multicolumn{1}{c}{81.0} & 81.7 & \underline{85.0} & \underline{86.3} & \underline{77.4} & 74.3  & 80.2 & 79.9 & 80.4 & 79.9 & 73.6 & 72.8 & 86.8 & 87.0 & \underline{89.7} & \underline{89.1} & 83.9 & \textbf{80.5} \\
\midrule
\multicolumn{11}{l}{\textbf{Original + Domain transfer (\textbf{ours})}} \\
\phantom{-} \textsc{ttest+tran.} &
\underline{81.7} & 81.9 & 84.1 & 84.1 & 72.3 & 69.6 & \underline{81.3} & \underline{80.3} & \underline{83.3} & 81.0 & 72.4 & 69.7 & \underline{87.1} & \textbf{87.1} & 87.2 & 84.5 & 72.7 & 69.7 \\
\midrule
\multicolumn{11}{l}{\textbf{Original + Summarization (\textbf{ours})}} \\
\phantom{-} \textsc{ttest+sum.} & \textbf{86.2} & \textbf{84.7} & \textbf{91.6} & \textbf{88.8} & 76.6 & 74.0  & 81.1 & \textbf{81.2} & \textbf{87.3} & \textbf{84.3} & 74.3 & \underline{73.8} & \textbf{87.8} & 86.8 & \textbf{92.8} & \textbf{90.2} & 83.0 & 75.9 \\
\bottomrule
\end{tabular}
\caption{
\textbf{Out-of-distribution generalization with data augmentation.} \emph{Original only}: baseline model trained solely on IMDb reviews, without CAD or data augmentation. \emph{+CAD}: augments IMDb training samples with manually constructed counterfactuals. \emph{+CORE}: augments training samples with automatically generated counterfactuals. \emph{+Domain transfer} and \emph{+Summarization} augment the training data with our newly proposed strategies. \textbf{Best} model in bold with the \underline{runner-up} underlined.}
\label{table:RQ2}
\end{table*}

\paragraph{Impact of the pre-trained multilingual models:}
We compare the OOD generalization of LaBSE, mBERT, and XLM-R.
The results in \cref{table:RQ1} show XLM-R as the top performer, consistently surpassing both LaBSE and mBERT. Despite having only 768 trainable parameters (frozen encoder with trainable logistic regression layer) against mBERT's 110M (fully tuned), it is surprising that LaBSE is at least on par with mBERT on non-English OOD data, except for \emph{translate-test}. This suggests a stronger bias towards English in mBERT compared to LaBSE, also evidenced by an 8.7\% drop in mBERT's ID \emph{zero-shot} performance between English and non-English, whereas this difference is just 0.1\% for LaBSE.

\paragraph{Impact of the transfer strategies:}
We assess the \emph{translate-train} and \emph{translate-test} strategies for OOD generalization against the \emph{zero-shot} approach.
The results in \cref{table:RQ1} reveal large OOD generalization gains for non-English languages using \emph{translate-test} and mBERT, with accuracy gains between $+$5.6\% and $+$9.3\%. This supports our previous discussion of mBERT being more biased towards English. For LaBSE, \emph{translate-train} is most effective on \emph{Amazon} and \emph{Restaurants}, with average accuracy boosts of $+$2.1\% and $+$2.3\% respectively, but not for \emph{Tweets} ($-$1.8\%).
For XLM-R, \emph{Restaurants} and \emph{Tweets} benefit most from translation: \emph{translate-train} (\emph{translate-test}) surpass \emph{zero-shot} with respective gains of $+$3.8\% ($+$2.3\%) and $+$3.3\% ($+$2.5\%). 
In conclusion, while translation-based strategies can further boost the OOD generalization zero-shot cross-lingual transfer, the benefits are dependent on the multilingual model and OOD test data.

\subsection{Out-of-distribution generalization with data augmentation}\label{subsec:results_rq2}

To address 
\textbf{(RQ2)} on achieving better OOD generalization,
we first analyze the effect of augmenting training data with the manually constructed counterfactuals of \citet{kaushik2019learning}. These counterfactuals will serve as an upper baseline against which we will subsequently compare the performance of models trained on 
\begin{enumerate*}[(i)]
    \item counterfactuals generated by the state-of-the-art in automatic counterfactual construction, \ie CORE  \cite{dixit-etal-2022-core}, and
    \item our LLM \emph{domain transferred} and \emph{summarized} augmented data. 
\end{enumerate*}

\paragraph{Manually constructed counterfactuals:} Comparing the \emph{original + CAD} results in \cref{table:RQ2} to the corresponding \emph{original only} results, reveals that augmenting training data with CAD consistently boosts OOD generalization, across all datasets and both for English and non-English test samples.
Accuracy gains averaged over the non-English languages for OOD vary between 7\%--14.8\%, 1.2\%--4.7\%, and 0.4\%--7.9\% for respectively LaBSE, mBERT, and XLM-R. This confirms that the English OOD generalization gains of CAD based training \cite{kaushik2019learning} translate well to non-English OOD test data in a 
cross-lingual setting.

\paragraph{Impact of LLM-based data augmentation on cross-lingual OOD generalization:} 
As an alternative to costly manually constructed counterfactuals, we investigate the viability of \emph{automatic} data augmentation: CORE from \citet{dixit-etal-2022-core} (replacing humans with the LLM for counterfactual creation), as well as our newly proposed \emph{domain transfer} and \emph{summarization} strategies described in \cref{subsec:data_augmentation}.
First, we compare the non-English OOD generalization of models trained with augmented data to models trained solely on original data. 
\cref{table:RQ2} shows clear non-English OOD improvements for 
all of LaBSE, mBERT, and XLM-R, with respective gains over \emph{original only} ranging from:
\begin{enumerate*}[(i)]
    \item 3.3\%--14.1\%, 0\%--2.1\%, and 1.4\%--8.8\% for CORE,
    \item 0.8\%--14.3\%, $-$1.4\%--2.5\%, and $-$2.0\%--3\% for \emph{domain transfer}, and
    \item 5.2\%--17.1\%, 2.7\%--5.4\%, and 1.3\%--8.7\% for \emph{summarization}.
\end{enumerate*}
The drops $-$1.4\% and $-$2.0\% for mBERT and XLM-R on \emph{Tweets} suggest that \emph{domain transfer} is less effective when the discrepancy between test and training domains is excessively large: the IMDb training data, similar to the \emph{Amazon} and \emph{Restaurant} domains, comprises reviews, whereas \emph{Tweets} do not.

The bold and underlined scores in \cref{table:RQ2} denote the top two results. Our \emph{summarization} strategy achieves the best non-English OOD generalization on \emph{Amazon} and \emph{Restaurants}, on par with  (or surpassing) models trained on CAD. On \emph{Tweets}, while \emph{summarization} still improves models trained solely on the original data, training on CAD or CORE (XLM-R) yields the best results. 

These findings support the efficacy of cost-effective data augmentation as a viable alternative to manually constructed counterfactuals for non-English test data. It is worth noting that our \emph{summarization} and \emph{domain transfer} methods scale linearly, only requiring a single transformation of training samples for each class. However, it is doubtful that
CAD and CORE can be similarly expanded beyond binary sentiment classification due to their quadratic data complexity: counterfactuals have to be constructed among every pair of classes.

\paragraph{Impact of LLM-based data augmentation on mono-lingual OOD generalization:}
Thus far, our analysis has primarily focused on the generalization from English ID training data to non-English OOD test data. Here, we investigate whether our \emph{summarization} and \emph{domain transfer} strategies can also help classifiers generalize in the well-studied monolingual setup, \ie from English training data to English OOD test data. In this setup, the \emph{translate-test} step is omitted: both the English ID training reviews from IMDb and the English OOD test samples are summarized or domain transferred, without any prior translation.

Comparing the \en\ scores across the different transfer strategies in \cref{table:RQ2} for each of LaBSE, mBERT, and XLM-R, reveals findings similar to the OOD generalization to non-English languages.
\begin{enumerate*}[(i)]
    \item For \emph{Amazon} and \emph{Restaurants}, all data augmentation approaches deliver classifiers that better generalize OOD compared to the \emph{original only} classifiers trained without augmented data. Our \emph{summarization} strategy achieves the best overall results, surpassing both classifiers trained on CORE and manually constructed counterfactuals (CAD), except for mBERT and \emph{Amazon}, where CAD results in a minor accuracy gain of 0.6\% over \emph{summarization}.
    \item  Surprisingly, for \emph{Tweets}, only classifiers trained on manually constructed CAD show consistent OOD generalization improvements over \emph{original-only} classifiers. This is in contrast to the results observed for non-English, where CORE and our \emph{summarization} augmentation approach were able to improve upon the \emph{original-only} classifiers.
\end{enumerate*}

Overall, these results highlight that our \emph{summarization} strategy can also benefit monolingual OOD generalization, surpassing classifiers augmented either with CAD or CORE generated counterfactuals for \emph{Amazon} and \emph{Restaurants}.

\begin{table}[t!]
\smaller
\centering
\addtolength{\tabcolsep}{-3.3pt}
\begin{tabular}{l cc cc cc}
\toprule
& \multicolumn{2}{c}{\textsc{amazon}} & \multicolumn{2}{c}{\textsc{restaurants}} & \multicolumn{2}{c}{\textsc{tweets}}\\
\cmidrule(lr){2-3} \cmidrule(lr){4-5} \cmidrule(lr){6-7}  
Method & \en & \textsc{non-en} & \en & \textsc{non-en} &  \en & \textsc{non-en} \\ \midrule
\multicolumn{7}{l}{\textbf{LaBSE}} \\
\textsc{zshot}  & 77.1 & 79.7 & 83.6 & 83.7 & 81.9 & 71.8 \\
\phantom{-} \textsc{+ttest}  & 77.1 & 78.9 & 83.6 & 84.0 & \textbf{81.9} & 73.1 \\
\phantom{-} \textsc{+sum.} & \textbf{86.2} & \textbf{84.7} & \textbf{91.6} & \textbf{88.8} & 76.6 & \textbf{74.0} \\
\midrule
\multicolumn{7}{l}{\textbf{mBERT}} \\
\textsc{zshot}  & 80.7 & 73.6 & 82.4 & 72.5 & 77.8 & 63.5  \\
\phantom{-} \textsc{+ttest}  & 80.7 & 79.6 & 82.4 & 80.0 & \textbf{77.8} & 72.0 \\
\phantom{-} \textsc{+sum.}  & \textbf{81.0} & \textbf{81.2} & \textbf{87.3} & \textbf{84.3} & 74.3 & \textbf{73.8} \\
\midrule
\multicolumn{7}{l}{\textbf{XLM-R}} \\
\textsc{zshot} & 87.8 & 87.7 & 89.4 & 84.9 & 86.3 & 75.1  \\
\phantom{-} \textsc{+ttest} & 87.8 & \textbf{88.0} & 89.4 & 87.1 & \textbf{86.3} & \textbf{77.8} \\
\phantom{-} \textsc{+sum.} & 87.8 & 86.8 & \textbf{92.8} & \textbf{90.2} & 83.0 & 75.9 \\
\bottomrule
\end{tabular}
\caption{
\textbf{Ablations} of our best data augmentation strategy: \emph{summarization}. \textsc{zshot}: trains on the original English and summarized English IMDb reviews. \textsc{+ttest}: additionally translates test samples to English. \textsc{+sum.}: further summarizes the English translated test samples prior inference.}
\label{table:ablations}
\end{table}

\paragraph{Ablations:}
We provide ablations in \cref{table:ablations} for our most effective strategy, \ie \emph{summarization}, and find that:
\begin{enumerate}[(i), wide,labelindent=0pt,nosep]
    \item The benefits of translating test samples into English (\emph{translate-test}) versus solely augmenting the training data with summaries (\emph{zero-shot}) vary based on the multilingual and/or OOD test data: there are clear OOD improvements to non-English samples for mBERT and XLM-R, but results for LaBSE are mixed and 
    comparable to the \emph{zero-shot} strategy;
    \item More importantly, further summarizing the English translated test samples improves OOD generalization more than solely translating them to English, consistently boosting accuracies by
    up to $+$5\% for LaBSE and $+$4.3\% mBERT, across all datasets. For XLM-R, summarization slightly reduces accuracy, \eg $-$1.2\% for non-English languages on \emph{Amazon} and $-$1.9\% for \emph{Tweets} compared to translation alone, yet still boosts OOD generalization to \emph{Restaurants} by 3.1\% over \emph{translate-test}.
\end{enumerate}

\paragraph{Cost-effectiveness of LLM-based augmentation:}
To assess the cost-effectiveness of our LLM-based augmentation, we discuss the costs of our best approach, \ie \emph{summarization}, and compare it to that cost of employing human workers to manually construct counterfactuals. \citet{kaushik2019learning} report that human workers spent an average of 5 minutes revising a single IMDb review, with each worker earning \$0.65 per revised review. Therefore, manually revising 1.7K training reviews incurs a total cost of $\approx$\$1,105 and $\approx$141 hours of labor.

In contrast, our summarization strategy costs \$0.0003 on average for summarizing a single training IMDb review, totaling \$0.51 for all 1.7K training reviews. However, our best OOD generalization is achieved not only by summarizing training reviews, but also by using an LLM during inference to:
\begin{enumerate*}[(1)]
    \item \label{it:translate} translate non-English test samples to English (\emph{translate-test}), and
    \item \label{it:summarize} further summarize the English translated test samples.
\end{enumerate*}
For \ref{it:translate}, the cost is \$0.00015 per OOD sample. For \ref{it:summarize}, an additional cost of \$0.00007 is required per OOD sample.\footnote{Summarizing OOD test samples is less costly than summarizing IMDb training samples due to the test samples comprising fewer tokens.} The reported costs per test sample are taken as the average among all OOD test sets and non-English languages.

In conclusion, our \emph{summarization} strategy costs \$0.51 to summarize all 1.7K training samples, and \$0.00022 (=\ref{it:translate}+\ref{it:summarize}) per inference. Thus, for the same cost of employing human workers for CAD creation ($\approx$ \$1,105), our \emph{summarization} strategy enables inference for 5M test samples. 
Note, however, that the best overall performance of classifiers augmented with CAD are achieved for \emph{translate-test}. Therefore, if we also account for translation costs of the CAD-augmented classifiers, our \emph{summarization} method can perform inference for 15M test samples for the same cost as employing human workers for CAD creation.
This demonstrates the cost-effectiveness of our \emph{summarization} approach when scaled up to 5M test samples as compared to \emph{zero-shot} +CAD, and up to 15M when compared to \emph{translate-test} +CAD. For future work, exploring open-source LLMs -or translation and summarization models could prove valuable for reducing inference costs.

\section{Conclusions}
\label{sec:conclusions}
We explored the generalization of zero-shot cross-lingual transfer to out-of-distribution (OOD) test data, considering both \emph{language} and \emph{domain} shifts.
Our experiments on binary sentiment classification with pre-trained multilingual models LaBSE, mBERT, and XLM-R finetuned on English IMDb movie reviews and evaluated on non-English test samples comprising \emph{Amazon} product reviews, \emph{Restaurant} feedback, and \emph{Tweets}, demonstrate that model performance substantially degrades, aligning with previous OOD generalization studies in a monolingual English setting. We also found that mBERT and XLM-R suffer more from performance reduction on OOD in non-English languages compared to English OOD degradation,
while LaBSE's generalization strongly depends on the OOD dataset.
Our experiments with models finetuned on original data augmented with manually constructed English counterfactual (CAD) IMDb reviews show that CAD's OOD generalization gains observed in a monolingual English setting also translate well to a zero-shot cross-lingual setup.
Finally, to avoid costly manually constructed counterfactuals, we propose two new data augmentation approaches for OOD generalization based on large language models: 
\begin{enumerate*}[(i)]
    \item \emph{domain transfer}, and
    \item \emph{summarization}
\end{enumerate*}.
Models trained with data augmented by our \emph{summarization} strategy, show substantial gains across all datasets and models, and on \emph{Amazon} and \emph{Restaurants} surpassing models either augmented with
\begin{enumerate*}[(i)]
    \item manually constructed CAD \cite{kaushik2019learning}, or
    \item state-of-the-art generated CORE counterfactuals \cite{dixit-etal-2022-core}.
\end{enumerate*}

\section*{Limitations}

\paragraph{Task domain:}
In this exploratory study, we only presented results for zero-shot cross-lingual binary sentiment classification. To investigate whether our findings generalize beyond binary classification, and to other non-classification tasks, further analysis is required. Nevertheless, as mentioned in \cref{subsec:results_rq2}, our data augmentation approaches scale better for classification tasks with more than two classes, since it only requires summarizing/transferring the training samples of each class once, whereas it is unclear how to scale counterfactuals to a larger number of classes.

\paragraph{Automatically translated in-distribution test data:}
Since we followed a similar setup as \citet{kaushik2019learning}, our experiments used the IMDb movie reviews as in-distribution sentiment data.
While the main focus in our study is on out-of-distribution generalization, the in-distribution test set was only provided in English.
Hence, we used translation tools to automatically translate the English IMDb test set to the considered non-English languages.
This may have caused annotation artifacts in the translated in-distribution tests, making it unclear how well the reported in-distribution results for non-English languages match real-world test data for non-English languages.

\paragraph{Translate-test based on a multilingual model:}
As our aim was to analyze the out-of-distribution generalization of multilingual models and compare their performance, we did not include results for the \emph{translate-test} based on a monolingual English model.
We believe that using such a monolingual model could further boost the accuracy of \emph{translate-test}, as well as for our \emph{summarization} and \emph{domain transfer} strategies.
However, we leave exploration thereof for future work.

\paragraph{Applicability to low-resource languages:}
The effectiveness of the \emph{translate-test} and \emph{translate-train} approaches are highly dependent on the accuracy of the adopted machine translation system. In this study, we used \texttt{ChatGPT-turbo (v0301)} as our translation tool, and found it to produce high-quality translations for all languages considered in our experiments, \ie boosting OOD generalization compared to the \emph{zero-shot} strategy. However, such machine translations systems may not work well for low-resource languages that lack high-quality translation data.

\section*{Ethics Statement}
Since our data augmentation methods use LLMs to generate summaries or create domain-transferred training (and test) samples, any biases present in the data used to train these LLMs could be transferred to the augmented data. We should therefore be careful to ensure that these biases do not carry over when training models on the augmented data, to avoid models that could discriminate against and/or potentially be harmful to certain demographics.

\section*{Acknowledgements}
This work was funded in part by Flanders Innovation \& Enterpreneurship (VLAIO), through Baekeland project-HBC.2019.2221 in collaboration with Sinch Chatlayer; and in part by the Flemish government under the ``Onderzoeksprogramma Artifici\"ele Intelligentie (AI) Vlaanderen'' (AI Research Program).

\bibliography{anthology,custom}

\begin{thebibliography}{49}
\expandafter\ifx\csname natexlab\endcsname\relax\def\natexlab#1{#1}\fi

\bibitem[{Arora et~al.(2021)Arora, Huang, and He}]{arora-etal-2021-types}
Udit Arora, William Huang, and He~He. 2021.
\newblock \href {https://doi.org/10.18653/v1/2021.emnlp-main.835} {Types of
  out-of-distribution texts and how to detect them}.
\newblock In \emph{Proceedings of the 2021 Conference on Empirical Methods in
  Natural Language Processing}, pages 10687--10701, Online and Punta Cana,
  Dominican Republic. Association for Computational Linguistics.

\bibitem[{Artetxe et~al.(2023)Artetxe, Goswami, Bhosale, Fan, and
  Zettlemoyer}]{artetxe2023revisiting}
Mikel Artetxe, Vedanuj Goswami, Shruti Bhosale, Angela Fan, and Luke
  Zettlemoyer. 2023.
\newblock Revisiting machine translation for cross-lingual classification.
\newblock \emph{arXiv preprint arXiv:2305.14240}.

\bibitem[{Artetxe et~al.(2020{\natexlab{a}})Artetxe, Labaka, and
  Agirre}]{artetxe-etal-2020-translation}
Mikel Artetxe, Gorka Labaka, and Eneko Agirre. 2020{\natexlab{a}}.
\newblock \href {https://doi.org/10.18653/v1/2020.emnlp-main.618} {Translation
  artifacts in cross-lingual transfer learning}.
\newblock In \emph{Proceedings of the 2020 Conference on Empirical Methods in
  Natural Language Processing (EMNLP)}, pages 7674--7684, Online. Association
  for Computational Linguistics.

\bibitem[{Artetxe et~al.(2020{\natexlab{b}})Artetxe, Ruder, and
  Yogatama}]{artetxe-etal-2020-cross}
Mikel Artetxe, Sebastian Ruder, and Dani Yogatama. 2020{\natexlab{b}}.
\newblock \href {https://doi.org/10.18653/v1/2020.acl-main.421} {On the
  cross-lingual transferability of monolingual representations}.
\newblock In \emph{Proceedings of the 58th Annual Meeting of the Association
  for Computational Linguistics}, pages 4623--4637, Online. Association for
  Computational Linguistics.

\bibitem[{Artetxe and Schwenk(2019)}]{artetxe-schwenk-2019-massively}
Mikel Artetxe and Holger Schwenk. 2019.
\newblock \href {https://doi.org/10.1162/tacl_a_00288} {Massively multilingual
  sentence embeddings for zero-shot cross-lingual transfer and beyond}.
\newblock \emph{Transactions of the Association for Computational Linguistics},
  7:597--610.

\bibitem[{Barbieri et~al.(2022)Barbieri, Espinosa~Anke, and
  Camacho-Collados}]{barbieri-etal-2022-xlm}
Francesco Barbieri, Luis Espinosa~Anke, and Jose Camacho-Collados. 2022.
\newblock \href {https://aclanthology.org/2022.lrec-1.27} {{XLM}-{T}:
  Multilingual language models in {T}witter for sentiment analysis and beyond}.
\newblock In \emph{Proceedings of the Thirteenth Language Resources and
  Evaluation Conference}, pages 258--266, Marseille, France. European Language
  Resources Association.

\bibitem[{Chi et~al.(2022)Chi, Huang, Dong, Ma, Zheng, Singhal, Bajaj, Song,
  Mao, Huang, and Wei}]{chi-etal-2022-xlm}
Zewen Chi, Shaohan Huang, Li~Dong, Shuming Ma, Bo~Zheng, Saksham Singhal, Payal
  Bajaj, Xia Song, Xian-Ling Mao, Heyan Huang, and Furu Wei. 2022.
\newblock \href {https://doi.org/10.18653/v1/2022.acl-long.427} {{XLM}-{E}:
  Cross-lingual language model pre-training via {ELECTRA}}.
\newblock In \emph{Proceedings of the 60th Annual Meeting of the Association
  for Computational Linguistics (Volume 1: Long Papers)}, pages 6170--6182,
  Dublin, Ireland. Association for Computational Linguistics.

\bibitem[{Conneau et~al.(2020)Conneau, Khandelwal, Goyal, Chaudhary, Wenzek,
  Guzm{\'a}n, Grave, Ott, Zettlemoyer, and
  Stoyanov}]{conneau-etal-2020-unsupervised}
Alexis Conneau, Kartikay Khandelwal, Naman Goyal, Vishrav Chaudhary, Guillaume
  Wenzek, Francisco Guzm{\'a}n, Edouard Grave, Myle Ott, Luke Zettlemoyer, and
  Veselin Stoyanov. 2020.
\newblock \href {https://doi.org/10.18653/v1/2020.acl-main.747} {Unsupervised
  cross-lingual representation learning at scale}.
\newblock In \emph{Proceedings of the 58th Annual Meeting of the Association
  for Computational Linguistics}, pages 8440--8451, Online. Association for
  Computational Linguistics.

\bibitem[{Conneau and Lample(2019)}]{conneau2019cross}
Alexis Conneau and Guillaume Lample. 2019.
\newblock Cross-lingual language model pretraining.
\newblock \emph{Advances in neural information processing systems}, 32.

\bibitem[{Cooper~Stickland et~al.(2023)Cooper~Stickland, Sengupta, Krone,
  Mansour, and He}]{cooper-stickland-etal-2023-robustification}
Asa Cooper~Stickland, Sailik Sengupta, Jason Krone, Saab Mansour, and He~He.
  2023.
\newblock \href {https://aclanthology.org/2023.eacl-main.100} {Robustification
  of multilingual language models to real-world noise in crosslingual zero-shot
  settings with robust contrastive pretraining}.
\newblock In \emph{Proceedings of the 17th Conference of the European Chapter
  of the Association for Computational Linguistics}, pages 1375--1391,
  Dubrovnik, Croatia. Association for Computational Linguistics.

\bibitem[{Costa-juss{\`a} et~al.(2022)Costa-juss{\`a}, Cross, {\c{C}}elebi,
  Elbayad, Heafield, Heffernan, Kalbassi, Lam, Licht, Maillard
  et~al.}]{costa2022no}
Marta~R Costa-juss{\`a}, James Cross, Onur {\c{C}}elebi, Maha Elbayad, Kenneth
  Heafield, Kevin Heffernan, Elahe Kalbassi, Janice Lam, Daniel Licht, Jean
  Maillard, et~al. 2022.
\newblock No language left behind: Scaling human-centered machine translation.
\newblock \emph{arXiv preprint arXiv:2207.04672}.

\bibitem[{De~Raedt et~al.(2022)De~Raedt, Godin, Develder, and
  Demeester}]{de-raedt-etal-2022-robustifying}
Maarten De~Raedt, Fr{\'e}deric Godin, Chris Develder, and Thomas Demeester.
  2022.
\newblock \href {https://aclanthology.org/2022.emnlp-main.783} {Robustifying
  sentiment classification by maximally exploiting few counterfactuals}.
\newblock In \emph{Proceedings of the 2022 Conference on Empirical Methods in
  Natural Language Processing}, pages 11386--11400, Abu Dhabi, United Arab
  Emirates. Association for Computational Linguistics.

\bibitem[{Devlin et~al.(2019)Devlin, Chang, Lee, and
  Toutanova}]{devlin-etal-2019-bert}
Jacob Devlin, Ming-Wei Chang, Kenton Lee, and Kristina Toutanova. 2019.
\newblock \href {https://doi.org/10.18653/v1/N19-1423} {{BERT}: Pre-training of
  deep bidirectional transformers for language understanding}.
\newblock In \emph{Proceedings of the 2019 Conference of the North {A}merican
  Chapter of the Association for Computational Linguistics: Human Language
  Technologies, Volume 1 (Long and Short Papers)}, pages 4171--4186,
  Minneapolis, Minnesota. Association for Computational Linguistics.

\bibitem[{Dixit et~al.(2022)Dixit, Paranjape, Hajishirzi, and
  Zettlemoyer}]{dixit-etal-2022-core}
Tanay Dixit, Bhargavi Paranjape, Hannaneh Hajishirzi, and Luke Zettlemoyer.
  2022.
\newblock \href {https://aclanthology.org/2022.findings-emnlp.216} {{CORE}: A
  retrieve-then-edit framework for counterfactual data generation}.
\newblock In \emph{Findings of the Association for Computational Linguistics:
  EMNLP 2022}, pages 2964--2984, Abu Dhabi, United Arab Emirates. Association
  for Computational Linguistics.

\bibitem[{Duh et~al.(2011)Duh, Fujino, and Nagata}]{duh-etal-2011-machine}
Kevin Duh, Akinori Fujino, and Masaaki Nagata. 2011.
\newblock \href {https://aclanthology.org/P11-2075} {Is machine translation
  ripe for cross-lingual sentiment classification?}
\newblock In \emph{Proceedings of the 49th Annual Meeting of the Association
  for Computational Linguistics: Human Language Technologies}, pages 429--433,
  Portland, Oregon, USA. Association for Computational Linguistics.

\bibitem[{Feng et~al.(2022)Feng, Yang, Cer, Arivazhagan, and
  Wang}]{feng-etal-2022-language}
Fangxiaoyu Feng, Yinfei Yang, Daniel Cer, Naveen Arivazhagan, and Wei Wang.
  2022.
\newblock \href {https://doi.org/10.18653/v1/2022.acl-long.62}
  {Language-agnostic {BERT} sentence embedding}.
\newblock In \emph{Proceedings of the 60th Annual Meeting of the Association
  for Computational Linguistics (Volume 1: Long Papers)}, pages 878--891,
  Dublin, Ireland. Association for Computational Linguistics.

\bibitem[{Gardner et~al.(2020)Gardner, Artzi, Basmov, Berant, Bogin, Chen,
  Dasigi, Dua, Elazar, Gottumukkala, Gupta, Hajishirzi, Ilharco, Khashabi, Lin,
  Liu, Liu, Mulcaire, Ning, Singh, Smith, Subramanian, Tsarfaty, Wallace,
  Zhang, and Zhou}]{gardner-etal-2020-evaluating}
Matt Gardner, Yoav Artzi, Victoria Basmov, Jonathan Berant, Ben Bogin, Sihao
  Chen, Pradeep Dasigi, Dheeru Dua, Yanai Elazar, Ananth Gottumukkala, Nitish
  Gupta, Hannaneh Hajishirzi, Gabriel Ilharco, Daniel Khashabi, Kevin Lin,
  Jiangming Liu, Nelson~F. Liu, Phoebe Mulcaire, Qiang Ning, Sameer Singh,
  Noah~A. Smith, Sanjay Subramanian, Reut Tsarfaty, Eric Wallace, Ally Zhang,
  and Ben Zhou. 2020.
\newblock \href {https://doi.org/10.18653/v1/2020.findings-emnlp.117}
  {Evaluating models{'} local decision boundaries via contrast sets}.
\newblock In \emph{Findings of the Association for Computational Linguistics:
  EMNLP 2020}, pages 1307--1323, Online. Association for Computational
  Linguistics.

\bibitem[{Gururangan et~al.(2018)Gururangan, Swayamdipta, Levy, Schwartz,
  Bowman, and Smith}]{gururangan-etal-2018-annotation}
Suchin Gururangan, Swabha Swayamdipta, Omer Levy, Roy Schwartz, Samuel Bowman,
  and Noah~A. Smith. 2018.
\newblock \href {https://doi.org/10.18653/v1/N18-2017} {Annotation artifacts in
  natural language inference data}.
\newblock In \emph{Proceedings of the 2018 Conference of the North {A}merican
  Chapter of the Association for Computational Linguistics: Human Language
  Technologies, Volume 2 (Short Papers)}, pages 107--112, New Orleans,
  Louisiana. Association for Computational Linguistics.

\bibitem[{Hendy et~al.(2023)Hendy, Abdelrehim, Sharaf, Raunak, Gabr,
  Matsushita, Kim, Afify, and Awadalla}]{hendy2023good}
Amr Hendy, Mohamed Abdelrehim, Amr Sharaf, Vikas Raunak, Mohamed Gabr, Hitokazu
  Matsushita, Young~Jin Kim, Mohamed Afify, and Hany~Hassan Awadalla. 2023.
\newblock How good are {GPT} models at machine translation? a comprehensive
  evaluation.
\newblock \emph{arXiv preprint arXiv:2302.09210}.

\bibitem[{Howard et~al.(2022)Howard, Singer, Lal, Choi, and
  Swayamdipta}]{howard-etal-2022-neurocounterfactuals}
Phillip Howard, Gadi Singer, Vasudev Lal, Yejin Choi, and Swabha Swayamdipta.
  2022.
\newblock \href {https://aclanthology.org/2022.findings-emnlp.371}
  {{N}euro{C}ounterfactuals: Beyond minimal-edit counterfactuals for richer
  data augmentation}.
\newblock In \emph{Findings of the Association for Computational Linguistics:
  EMNLP 2022}, pages 5056--5072, Abu Dhabi, United Arab Emirates. Association
  for Computational Linguistics.

\bibitem[{Hu et~al.(2020)Hu, Ruder, Siddhant, Neubig, Firat, and
  Johnson}]{hu2020xtreme}
Junjie Hu, Sebastian Ruder, Aditya Siddhant, Graham Neubig, Orhan Firat, and
  Melvin Johnson. 2020.
\newblock Xtreme: A massively multilingual multi-task benchmark for evaluating
  cross-lingual generalisation.
\newblock In \emph{International Conference on Machine Learning}, pages
  4411--4421. PMLR.

\bibitem[{Jiao et~al.(2023)Jiao, Wang, Huang, Wang, and Tu}]{jiao2023chatgpt}
Wenxiang Jiao, Wenxuan Wang, Jen-tse Huang, Xing Wang, and Zhaopeng Tu. 2023.
\newblock Is {ChatGPT} a good translator? a preliminary study.
\newblock \emph{arXiv preprint arXiv:2301.08745}.

\bibitem[{Joshi et~al.(2022)Joshi, Pan, and He}]{joshi-etal-2022-spurious}
Nitish Joshi, Xiang Pan, and He~He. 2022.
\newblock \href {https://aclanthology.org/2022.emnlp-main.666} {Are all
  spurious features in natural language alike? an analysis through a causal
  lens}.
\newblock In \emph{Proceedings of the 2022 Conference on Empirical Methods in
  Natural Language Processing}, pages 9804--9817, Abu Dhabi, United Arab
  Emirates. Association for Computational Linguistics.

\bibitem[{Katakkar et~al.(2022)Katakkar, Yoo, Wang, Lipton, and
  Kaushik}]{katakkar-etal-2022-practical}
Anurag Katakkar, Clay~H. Yoo, Weiqin Wang, Zachary Lipton, and Divyansh
  Kaushik. 2022.
\newblock \href {https://aclanthology.org/2022.blackboxnlp-1.29} {Practical
  benefits of feature feedback under distribution shift}.
\newblock In \emph{Proceedings of the Fifth BlackboxNLP Workshop on Analyzing
  and Interpreting Neural Networks for NLP}, pages 346--355, Abu Dhabi, United
  Arab Emirates (Hybrid). Association for Computational Linguistics.

\bibitem[{Kaushik et~al.(2019)Kaushik, Hovy, and Lipton}]{kaushik2019learning}
Divyansh Kaushik, Eduard Hovy, and Zachary Lipton. 2019.
\newblock Learning the difference that makes a difference with
  counterfactually-augmented data.
\newblock In \emph{International Conference on Learning Representations}.

\bibitem[{Kaushik et~al.(2020)Kaushik, Setlur, Hovy, and
  Lipton}]{kaushik2020explaining}
Divyansh Kaushik, Amrith Setlur, Eduard~H Hovy, and Zachary~Chase Lipton. 2020.
\newblock Explaining the efficacy of counterfactually augmented data.
\newblock In \emph{International Conference on Learning Representations}.

\bibitem[{Keung et~al.(2020)Keung, Lu, Szarvas, and
  Smith}]{keung-etal-2020-multilingual}
Phillip Keung, Yichao Lu, Gy{\"o}rgy Szarvas, and Noah~A. Smith. 2020.
\newblock \href {https://doi.org/10.18653/v1/2020.emnlp-main.369} {The
  multilingual {A}mazon reviews corpus}.
\newblock In \emph{Proceedings of the 2020 Conference on Empirical Methods in
  Natural Language Processing (EMNLP)}, pages 4563--4568, Online. Association
  for Computational Linguistics.

\bibitem[{Lauscher et~al.(2020)Lauscher, Ravishankar, Vuli{\'c}, and
  Glava{\v{s}}}]{lauscher-etal-2020-zero}
Anne Lauscher, Vinit Ravishankar, Ivan Vuli{\'c}, and Goran Glava{\v{s}}. 2020.
\newblock \href {https://doi.org/10.18653/v1/2020.emnlp-main.363} {From zero to
  hero: {O}n the limitations of zero-shot language transfer with multilingual
  {T}ransformers}.
\newblock In \emph{Proceedings of the 2020 Conference on Empirical Methods in
  Natural Language Processing (EMNLP)}, pages 4483--4499, Online. Association
  for Computational Linguistics.

\bibitem[{Liu and Nocedal(1989)}]{liu1989limited}
Dong~C Liu and Jorge Nocedal. 1989.
\newblock On the limited memory bfgs method for large scale optimization.
\newblock \emph{Mathematical programming}, 45(1-3):503--528.

\bibitem[{Liu et~al.(2021)Liu, Winata, Madotto, and
  Fung}]{liu-etal-2021-preserving}
Zihan Liu, Genta~Indra Winata, Andrea Madotto, and Pascale Fung. 2021.
\newblock \href {https://doi.org/10.18653/v1/2021.repl4nlp-1.8} {Preserving
  cross-linguality of pre-trained models via continual learning}.
\newblock In \emph{Proceedings of the 6th Workshop on Representation Learning
  for NLP (RepL4NLP-2021)}, pages 64--71, Online. Association for Computational
  Linguistics.

\bibitem[{Maas et~al.(2011)Maas, Daly, Pham, Huang, Ng, and
  Potts}]{maas-etal-2011-learning}
Andrew~L. Maas, Raymond~E. Daly, Peter~T. Pham, Dan Huang, Andrew~Y. Ng, and
  Christopher Potts. 2011.
\newblock \href {https://aclanthology.org/P11-1015} {Learning word vectors for
  sentiment analysis}.
\newblock In \emph{Proceedings of the 49th Annual Meeting of the Association
  for Computational Linguistics: Human Language Technologies}, pages 142--150,
  Portland, Oregon, USA. Association for Computational Linguistics.

\bibitem[{McCoy et~al.(2019)McCoy, Pavlick, and Linzen}]{mccoy-etal-2019-right}
Tom McCoy, Ellie Pavlick, and Tal Linzen. 2019.
\newblock \href {https://doi.org/10.18653/v1/P19-1334} {Right for the wrong
  reasons: Diagnosing syntactic heuristics in natural language inference}.
\newblock In \emph{Proceedings of the 57th Annual Meeting of the Association
  for Computational Linguistics}, pages 3428--3448, Florence, Italy.
  Association for Computational Linguistics.

\bibitem[{Ouyang et~al.(2022)Ouyang, Wu, Jiang, Almeida, Wainwright, Mishkin,
  Zhang, Agarwal, Slama, Gray et~al.}]{ouyangtraining}
Long Ouyang, Jeffrey Wu, Xu~Jiang, Diogo Almeida, Carroll Wainwright, Pamela
  Mishkin, Chong Zhang, Sandhini Agarwal, Katarina Slama, Alex Gray, et~al.
  2022.
\newblock Training language models to follow instructions with human feedback.
\newblock In \emph{Advances in Neural Information Processing Systems}.

\bibitem[{Pedregosa et~al.(2011)Pedregosa, Varoquaux, Gramfort, Michel,
  Thirion, Grisel, Blondel, Prettenhofer, Weiss, Dubourg
  et~al.}]{pedregosa2011scikit}
Fabian Pedregosa, Ga{\"e}l Varoquaux, Alexandre Gramfort, Vincent Michel,
  Bertrand Thirion, Olivier Grisel, Mathieu Blondel, Peter Prettenhofer, Ron
  Weiss, Vincent Dubourg, et~al. 2011.
\newblock Scikit-learn: Machine learning in {Python}.
\newblock \emph{The Journal Of Machine Learning Research}, 12:2825--2830.

\bibitem[{Peng et~al.(2023)Peng, Ding, Zhong, Shen, Liu, Zhang, Ouyang, and
  Tao}]{Peng2023ChatGPT4MT}
Keqin Peng, Liang Ding, Qihuang Zhong, Li~Shen, Xuebo Liu, Min Zhang, Yuanxin
  Ouyang, and Dacheng Tao. 2023.
\newblock \href {https://arxiv.org/abs/2303.13780} {Towards making the most of
  {ChatGPT} for machine translation}.
\newblock \emph{arxiv preprint}.

\bibitem[{Poliak et~al.(2018)Poliak, Naradowsky, Haldar, Rudinger, and
  Van~Durme}]{poliak-etal-2018-hypothesis}
Adam Poliak, Jason Naradowsky, Aparajita Haldar, Rachel Rudinger, and Benjamin
  Van~Durme. 2018.
\newblock \href {https://doi.org/10.18653/v1/S18-2023} {Hypothesis only
  baselines in natural language inference}.
\newblock In \emph{Proceedings of the Seventh Joint Conference on Lexical and
  Computational Semantics}, pages 180--191, New Orleans, Louisiana. Association
  for Computational Linguistics.

\bibitem[{Pontiki et~al.(2016)Pontiki, Galanis, Papageorgiou, Androutsopoulos,
  Manandhar, AL-Smadi, Al-Ayyoub, Zhao, Qin, De~Clercq, Hoste, Apidianaki,
  Tannier, Loukachevitch, Kotelnikov, Bel, Jim{\'e}nez-Zafra, and
  Eryi{\u{g}}it}]{pontiki-etal-2016-semeval}
Maria Pontiki, Dimitris Galanis, Haris Papageorgiou, Ion Androutsopoulos,
  Suresh Manandhar, Mohammad AL-Smadi, Mahmoud Al-Ayyoub, Yanyan Zhao, Bing
  Qin, Orph{\'e}e De~Clercq, V{\'e}ronique Hoste, Marianna Apidianaki, Xavier
  Tannier, Natalia Loukachevitch, Evgeniy Kotelnikov, Nuria Bel,
  Salud~Mar{\'\i}a Jim{\'e}nez-Zafra, and G{\"u}l{\c{s}}en Eryi{\u{g}}it. 2016.
\newblock \href {https://doi.org/10.18653/v1/S16-1002} {{S}em{E}val-2016 task
  5: Aspect based sentiment analysis}.
\newblock In \emph{Proceedings of the 10th International Workshop on Semantic
  Evaluation ({S}em{E}val-2016)}, pages 19--30, San Diego, California.
  Association for Computational Linguistics.

\bibitem[{Qui{\~n}onero-Candela et~al.(2008)Qui{\~n}onero-Candela, Sugiyama,
  Schwaighofer, and Lawrence}]{quinonero2008dataset}
Joaquin Qui{\~n}onero-Candela, Masashi Sugiyama, Anton Schwaighofer, and Neil~D
  Lawrence. 2008.
\newblock \emph{Dataset shift in machine learning}.
\newblock Mit Press.

\bibitem[{Ruder et~al.(2019)Ruder, Vuli{\'c}, and S{\o}gaard}]{ruder2019survey}
Sebastian Ruder, Ivan Vuli{\'c}, and Anders S{\o}gaard. 2019.
\newblock A survey of cross-lingual word embedding models.
\newblock \emph{Journal of Artificial Intelligence Research}, 65:569--631.

\bibitem[{Shi et~al.(2010)Shi, Mihalcea, and Tian}]{shi-etal-2010-cross}
Lei Shi, Rada Mihalcea, and Mingjun Tian. 2010.
\newblock \href {https://aclanthology.org/D10-1103} {Cross language text
  classification by model translation and semi-supervised learning}.
\newblock In \emph{Proceedings of the 2010 Conference on Empirical Methods in
  Natural Language Processing}, pages 1057--1067, Cambridge, MA. Association
  for Computational Linguistics.

\bibitem[{Wang et~al.(2021)Wang, Liu, Gui, Zhang, Zou, Zhou, Ye, Zhang, Zheng,
  Pang, Wu, Li, Zhang, Ma, Fei, Cai, Zhao, Hu, Yan, Tan, Hu, Bian, Liu, Qin,
  Zhu, Xing, Fu, Zhang, Peng, Zheng, Zhou, Wei, Qiu, and
  Huang}]{wang-etal-2021-textflint}
Xiao Wang, Qin Liu, Tao Gui, Qi~Zhang, Yicheng Zou, Xin Zhou, Jiacheng Ye,
  Yongxin Zhang, Rui Zheng, Zexiong Pang, Qinzhuo Wu, Zhengyan Li, Chong Zhang,
  Ruotian Ma, Zichu Fei, Ruijian Cai, Jun Zhao, Xingwu Hu, Zhiheng Yan, Yiding
  Tan, Yuan Hu, Qiyuan Bian, Zhihua Liu, Shan Qin, Bolin Zhu, Xiaoyu Xing,
  Jinlan Fu, Yue Zhang, Minlong Peng, Xiaoqing Zheng, Yaqian Zhou, Zhongyu Wei,
  Xipeng Qiu, and Xuanjing Huang. 2021.
\newblock \href {https://doi.org/10.18653/v1/2021.acl-demo.41} {{T}ext{F}lint:
  Unified multilingual robustness evaluation toolkit for natural language
  processing}.
\newblock In \emph{Proceedings of the 59th Annual Meeting of the Association
  for Computational Linguistics and the 11th International Joint Conference on
  Natural Language Processing: System Demonstrations}, pages 347--355, Online.
  Association for Computational Linguistics.

\bibitem[{Wang and Culotta(2020)}]{wang-culotta-2020-identifying}
Zhao Wang and Aron Culotta. 2020.
\newblock \href {https://doi.org/10.18653/v1/2020.findings-emnlp.308}
  {Identifying spurious correlations for robust text classification}.
\newblock In \emph{Findings of the Association for Computational Linguistics:
  EMNLP 2020}, pages 3431--3440, Online. Association for Computational
  Linguistics.

\bibitem[{Wang and Culotta(2021)}]{wang2021robustness}
Zhao Wang and Aron Culotta. 2021.
\newblock Robustness to spurious correlations in text classification via
  automatically generated counterfactuals.
\newblock In \emph{Proceedings of the AAAI Conference on Artificial
  Intelligence}, volume~35, pages 14024--14031.

\bibitem[{Wolf et~al.(2020)Wolf, Debut, Sanh, Chaumond, Delangue, Moi, Cistac,
  Rault, Louf, Funtowicz, Davison, Shleifer, von Platen, Ma, Jernite, Plu, Xu,
  Le~Scao, Gugger, Drame, Lhoest, and Rush}]{wolf-etal-2020-transformers}
Thomas Wolf, Lysandre Debut, Victor Sanh, Julien Chaumond, Clement Delangue,
  Anthony Moi, Pierric Cistac, Tim Rault, Remi Louf, Morgan Funtowicz, Joe
  Davison, Sam Shleifer, Patrick von Platen, Clara Ma, Yacine Jernite, Julien
  Plu, Canwen Xu, Teven Le~Scao, Sylvain Gugger, Mariama Drame, Quentin Lhoest,
  and Alexander Rush. 2020.
\newblock \href {https://doi.org/10.18653/v1/2020.emnlp-demos.6} {Transformers:
  State-of-the-art natural language processing}.
\newblock In \emph{Proceedings of the 2020 Conference on Empirical Methods in
  Natural Language Processing: System Demonstrations}, pages 38--45, Online.
  Association for Computational Linguistics.

\bibitem[{Xue et~al.(2021)Xue, Constant, Roberts, Kale, Al-Rfou, Siddhant,
  Barua, and Raffel}]{xue-etal-2021-mt5}
Linting Xue, Noah Constant, Adam Roberts, Mihir Kale, Rami Al-Rfou, Aditya
  Siddhant, Aditya Barua, and Colin Raffel. 2021.
\newblock \href {https://doi.org/10.18653/v1/2021.naacl-main.41} {m{T}5: A
  massively multilingual pre-trained text-to-text transformer}.
\newblock In \emph{Proceedings of the 2021 Conference of the North American
  Chapter of the Association for Computational Linguistics: Human Language
  Technologies}, pages 483--498, Online. Association for Computational
  Linguistics.

\bibitem[{Yanaka and Mineshima(2021)}]{yanaka-mineshima-2021-assessing}
Hitomi Yanaka and Koji Mineshima. 2021.
\newblock \href {https://doi.org/10.18653/v1/2021.blackboxnlp-1.26} {Assessing
  the generalization capacity of pre-trained language models through {J}apanese
  adversarial natural language inference}.
\newblock In \emph{Proceedings of the Fourth BlackboxNLP Workshop on Analyzing
  and Interpreting Neural Networks for NLP}, pages 337--349, Punta Cana,
  Dominican Republic. Association for Computational Linguistics.

\bibitem[{Yang et~al.(2021)Yang, Li, Cunningham, Zhang, Smyth, and
  Dong}]{yang-etal-2021-exploring}
Linyi Yang, Jiazheng Li, Padraig Cunningham, Yue Zhang, Barry Smyth, and Ruihai
  Dong. 2021.
\newblock \href {https://doi.org/10.18653/v1/2021.acl-long.26} {Exploring the
  efficacy of automatically generated counterfactuals for sentiment analysis}.
\newblock In \emph{Proceedings of the 59th Annual Meeting of the Association
  for Computational Linguistics and the 11th International Joint Conference on
  Natural Language Processing (Volume 1: Long Papers)}, pages 306--316, Online.
  Association for Computational Linguistics.

\bibitem[{Yu and Joty(2021)}]{yu-joty-2021-effective}
Tao Yu and Shafiq Joty. 2021.
\newblock \href {https://doi.org/10.18653/v1/2021.emnlp-main.668} {Effective
  fine-tuning methods for cross-lingual adaptation}.
\newblock In \emph{Proceedings of the 2021 Conference on Empirical Methods in
  Natural Language Processing}, pages 8492--8501, Online and Punta Cana,
  Dominican Republic. Association for Computational Linguistics.

\bibitem[{Zheng et~al.(2021)Zheng, Dong, Huang, Wang, Chi, Singhal, Che, Liu,
  Song, and Wei}]{zheng-etal-2021-consistency}
Bo~Zheng, Li~Dong, Shaohan Huang, Wenhui Wang, Zewen Chi, Saksham Singhal,
  Wanxiang Che, Ting Liu, Xia Song, and Furu Wei. 2021.
\newblock \href {https://doi.org/10.18653/v1/2021.acl-long.264} {Consistency
  regularization for cross-lingual fine-tuning}.
\newblock In \emph{Proceedings of the 59th Annual Meeting of the Association
  for Computational Linguistics and the 11th International Joint Conference on
  Natural Language Processing (Volume 1: Long Papers)}, pages 3403--3417,
  Online. Association for Computational Linguistics.

\end{thebibliography}
\bibliographystyle{acl_natbib}

\clearpage

\appendix

\section{Appendix}
\label{sec:appendix}

\begin{table}[t!]
\smaller
\centering
\addtolength{\tabcolsep}{0pt}
\begin{tabular}{l ccccccccccccc}
\toprule
\multicolumn{14}{c}{\#\,Test}   \\
\cmidrule(lr){2-14}
Dataset & \en & \de & \dutch & \fr & \es & \italian & \pt & \tu & \ru & \ja & \zh & \ar & \hi \\
\midrule
\textsc{amazon} & 4,000 & 4,000 & - & 4,000 & 4,000 & - & - & - & - & 4,000 & 4,000 & - & - \\
\textsc{tweets} & 580 & 580 & - & 580 & 580 & 580 & 580 & - & - & - & - & 580 & 580 \\
\textsc{restaurants} & 980 & - & 960 & 1,268 & 760 & - & - & 780 & 1,012 & - & - & - & - \\
\bottomrule
\end{tabular}
\caption{
\textbf{Out-of-distribution} dataset statistics.}
\label{table:OOD_datasets}
\end{table}

\begin{table}
\smaller
\centering
\addtolength{\tabcolsep}{7pt}
\begin{tabular}{lccc}
\toprule
\textsc{IMDB (en)} & \#\,Train & \#\,Val & \#\,Test  \\
\midrule
Original & 1,707 & 245 & 488 \\
CAD & 1,707 & 245 & - \\
\bottomrule
\end{tabular}
\addtolength{\tabcolsep}{-4pt}
\caption{
\textbf{In-distribution} dataset statistics.}
\label{table:ID_datasets}
\end{table}

\paragraph{Datasets:} \cref{table:ID_datasets,table:OOD_datasets} summarize respectively the number of \emph{out-of-distribution} test samples and the number of train, validation and test \emph{in-distribution} test samples. Note that the number of samples for \emph{translate-train} and \emph{translate-test} exactly match those shown in the tables. 

\paragraph{Prompts:} \cref{figure:translation_prompts,figure:augmentation_prompts} show our adopted prompts for instructing \texttt{ChatGPT-turbo} to translate 
\begin{enumerate*}[(i)]
    \item non-English out-of-distribution test samples into English for \emph{translate-test}, and
    \item English in-distribution English training and validation samples into non-English for \emph{translate-train}.
\end{enumerate*}

\paragraph{Detailed ID and OOD results per language:}
The in-distribution and out-of-distribution results per language are presented in \cref{table:ID_full,table:OOD_full}. As mentioned in \cref{subsec:results_rq1}, the \emph{translate-test} in-distribution scores  are not included for non-English languages. This is because these test sets are automatically translated versions of the original English test set. Including \emph{translate-test} scores would require translating the already translated test samples back to English, which would evaluate the quality of backtranslation rather than the \emph{translate-test} performance itself. In our pilot experiments, we observed that the backtranslation quality was quite high. As such, small differences in accuracy between the performance of \emph{translate-test} and the model performance on the original English test set appeared overly optimistic. Hence, we opted to exclude them.

\clearpage

\begin{figure*}
\centering
\begin{subfigure}{0.4\linewidth}
  \centering
  \begin{tikzpicture}
    \node[draw=mybluedark, rounded corners=2mm,text width=0.95\textwidth,inner sep=5mm, align=justify, line width=1pt] at (0,0) (box)
        {
        \vspace{0.2cm} \\ 
         Translate from \{Language\} to English. \\
        \{\textbf{Language}\}: \{test sample\} \\
        \textbf{English:}
        };
        \node[yshift=-0.4cm] at (box.north) {\textbf{Translate-test}};
  \end{tikzpicture}
\end{subfigure}
\hspace{0.6cm}
\begin{subfigure}{0.4\linewidth}
  \centering
  \begin{tikzpicture}
    \node[draw=mybluedark,rounded corners=2mm,text width=0.95\textwidth,inner sep=5mm, align=justify, line width=1pt] at (0,0) (box)
    {
    \vspace{0.2cm} \\
    Translate from English to \{Language\}. \\
    \textbf{English:} \{train sample\} \\
    \{\textbf{Language}\}:
    };
    \node[yshift=-0.4cm] at (box.north) {\textbf{Translate-train}};
  \end{tikzpicture}
\end{subfigure}
\caption{\textbf{Translation prompts} for \texttt{ChatGPT-turbo (v0301)}.}
\label{figure:translation_prompts}
\vspace{3cm}
\begin{subfigure}{0.4\linewidth}
  \centering
  \begin{tikzpicture}
    \node[draw=mybluedark,rounded corners=2mm,text width=1.0\textwidth,inner sep=5mm, align=justify, line width=1pt] at (0,0) (box)
        {
        \vspace{0.2cm} \\ 
         Summarize the review in a maximum of 10 words. \\
        \textbf{Review:} \{train -or English translated test sample\} \\
        };
        \node[yshift=-0.4cm] at (box.north) {\textbf{Summarization}};
  \end{tikzpicture}
\end{subfigure}
\hspace{1cm}
\begin{subfigure}{0.4\linewidth}
  \centering
  \begin{tikzpicture}
    \node[draw=mybluedark,rounded corners=2mm,text width=1.0\textwidth,inner sep=5mm, align=justify, line width=1pt] at (0,0) (box)
    {
    \vspace{0.2cm} \\
    Make minimal changes to adapt the review such that it becomes about books. \\
    \textbf{Review:} \{train or English-translated test sample\} \\
    };
    \node[yshift=-0.4cm] at (box.north) {\textbf{Domain transfer}};
  \end{tikzpicture}
\end{subfigure}
\caption{\textbf{Data augmentation prompts} for \texttt{ChatGPT-turbo (v0301)}. \textbf{Left:} \emph{Summarization} prompt. \textbf{Right:} \emph{Domain transfer} prompt.}
\label{figure:augmentation_prompts}
\end{figure*}

\begin{table*}
\begin{subtable}{\textwidth}
\smaller
\centering
\addtolength{\tabcolsep}{1pt}
\renewcommand{\arraystretch}{0.95}
\begin{tabular}{l ccccccc ccccccc}
\toprule
\multicolumn{14}{c}{\textbf{IMDB}} \\
\toprule
\toprule
\multicolumn{14}{c}{\textbf{LaBSE}} \\
\cmidrule(lr){2-14}
Method & \en & \de & \dutch & \fr & \es & \italian & \pt & \tu & \ru & \ar & \hi & \ja & \zh \\ \midrule 
\multicolumn{14}{l}{\textbf{Original only}} \\
- \textsc{zshot} & 85.0 & 85.3 & 86.0 & 85.9 & 86.1 & 85.4 & 85.5 & 83.5 & 85.1 & 85.2 & 81.2 & 83.5 & 86.0 \\
- \textsc{ttrain} & 85.0 & 86.0 & 87.0 & 84.5 & 87.1 & 85.4 & 86.9 & 83.0 & 86.5 & 85.0 & 81.8 & 83.9 & 85.6 \\
\midrule
\multicolumn{14}{l}{\textbf{Original \& CAD} \cite{kaushik2019learning}} \\
- \textsc{zshot} & 81.4 & 82.0 & 80.1 & 82.0 & 82.6 & 81.6 & 81.6 & 80.5 & 80.1 & 79.3 & 80.1 & 80.3 & 79.5 \\
- \textsc{ttrain} & 81.4 & 83.0 & 80.7 & 82.4 & 83.0 & 81.8 & 83.8 & 82.0 & 80.7 & 78.7 & 78.7 & 80.7 & 79.1 \\
\midrule
\multicolumn{14}{l}{\textbf{Original \& CORE} \cite{dixit-etal-2022-core}} \\
- \textsc{zshot} & 80.1 & 77.9 & 80.3 & 79.3 & 81.4 & 79.3 & 78.7 & 79.1 & 78.3 & 79.9 & 75.4 & 79.5 & 79.1 \\
\midrule
\multicolumn{14}{l}{\textbf{Domain transfer (\textbf{ours})}} \\
- \textsc{zshot}\textsuperscript{$\spadesuit$} & 83.3 & 84.5 & 84.5 & 84.4 & 86.0 & 85.4 & 85.5 & 82.3 & 83.8 & 84.6 & 79.0 & 82.7 & 83.3 \\
\phantom{-} \textsc{+trans.} & 85.5 & - & - & - & - & - & - & - & - & - & - & - & - \\
\midrule
\multicolumn{14}{l}{\textbf{Summarization (\textbf{ours})}} \\
- \textsc{zshot}\textsuperscript{$\spadesuit$} & 83.6 & 84.0 & 85.9 & 84.8 & 85.0 & 84.0 & 86.1 & 82.4 & 84.2 & 84.8 & 80.9 & 85.7 & 83.6 \\
\phantom{-} \textsc{+sum.} & 86.7 & - & - & - & - & - & - & - & - & - & - & - & - \\
\bottomrule
\end{tabular}
\label{table:ID_LaBSE}
\end{subtable}
\begin{subtable}{\textwidth}
\smaller
\centering
\addtolength{\tabcolsep}{1pt}
\renewcommand{\arraystretch}{0.95}
\begin{tabular}{l ccccccc ccccccc}
\toprule
\toprule
\multicolumn{14}{c}{\textbf{mBERT}} \\
\cmidrule(lr){2-14}
Method & \en & \de & \dutch & \fr & \es & \italian & \pt & \tu & \ru & \ar & \hi & \ja & \zh \\ \midrule 
\multicolumn{14}{l}{\textbf{Original only}} \\
- \textsc{zshot} & 89.5 & 84.0 & 77.8 & 84.2 & 86.9 & 83.4 & 83.2 & 76.1 & 80.0 & 75.2 & 72.2 & 81.9 & 84.8 \\
- \textsc{ttrain} & - & 87.2 & 89.1 & 89.1 & 90.2 & 88.7 & 88.8 & 87.4 & 87.8 & 84.1 & 81.9 & 87.1 & 88.5 \\
\midrule
\multicolumn{14}{l}{\textbf{Original \& CAD} \cite{kaushik2019learning}} \\
- \textsc{zshot} & 86.3 & 82.8 & 75.8 & 82.2 & 83.6 & 79.4 & 79.7 & 72.3 & 78.5 & 70.1 & 69.1 & 78.9 & 84.5 \\
- \textsc{ttrain} & - & 86.0 & 86.6 & 86.8 & 87.6 & 87.0 & 86.7 & 84.5 & 86.1 & 83.2 & 78.8 & 86.9 & 87.0 \\
\midrule
\multicolumn{14}{l}{\textbf{Original \& CORE} \cite{dixit-etal-2022-core}} \\
- \textsc{zshot} & 84.5 & 79.7 & 73.0 & 80.6 & 78.2 & 77.4 & 77.7 & 70.1 & 74.7 & 66.5 & 65.0 & 75.6 & 80.3 \\
\midrule
\multicolumn{14}{l}{\textbf{Domain transfer (\textbf{ours})}} \\
- \textsc{zshot}\textsuperscript{$\spadesuit$} & 86.7 & 82.9 & 76.7 & 84.1 & 84.3 & 82.0 & 82.0 & 75.8 & 77.7 & 74.3 & 71.1 & 79.3 & 84.4 \\
\phantom{-} \textsc{+trans.} & 87.8 & - & - & - & - & - & - & - & - & - & - & - & - \\
\midrule
\multicolumn{14}{l}{\textbf{Summarization (\textbf{ours})}} \\
- \textsc{zshot}\textsuperscript{$\spadesuit$} & 87.2 & 83.1 & 74.4 & 82.3 & 84.4 & 81.1 & 82.3 & 74.4 & 77.3 & 73.6 & 71.0 & 80.9 & 82.9 \\
\phantom{-} \textsc{+sum.} & 88.2 & - & - & - & - & - & - & - & - & - & - & - & - \\
\bottomrule
\end{tabular}
\end{subtable}
\begin{subtable}{\textwidth}
\smaller
\centering
\addtolength{\tabcolsep}{1pt}
\renewcommand{\arraystretch}{0.95}
\begin{tabular}{l ccccccc ccccccc}
\toprule
\toprule
\multicolumn{14}{c}{\textbf{XLM-R}} \\
\cmidrule(lr){2-14}
Method & \en & \de & \dutch & \fr & \es & \italian & \pt & \tu & \ru & \ar & \hi & \ja & \zh \\ \midrule 
\multicolumn{14}{l}{\textbf{Original only}} \\
- \textsc{zshot} & 92.4 & 90.4 & 90.9 & 89.9 & 89.8 & 89.5 & 90.7 & 88.5 & 89.4 & 84.7 & 82.3 & 85.4 & 89.6 \\
- \textsc{ttrain} & - & 91.4 & 92.2 & 91.7 & 91.6 & 91.3 & 91.8 & 91.0 & 90.9 & 89.2 & 86.4 & 89.2 & 91.1 \\
\midrule
\multicolumn{14}{l}{\textbf{Original \& CAD} \cite{kaushik2019learning}} \\
- \textsc{zshot} & 90.4 & 88.1 & 88.0 & 88.1 & 87.8 & 87.0 & 87.4 & 86.8 & 86.8 & 81.6 & 82.2 & 85.9 & 88.3 \\
- \textsc{ttrain} & - & 88.9 & 88.5 & 89.8 & 89.7 & 89.3 & 89.8 & 89.2 & 88.9 & 88.2 & 85.7 & 87.9 & 88.8 \\
\midrule
\multicolumn{14}{l}{\textbf{Original \& CORE} \cite{dixit-etal-2022-core}} \\
- \textsc{zshot} & 88.1 & 86.9 & 87.5 & 87.2 & 87.5 & 87.2 & 86.7 & 86.1 & 87.0 & 83.6 & 82.4 & 85.4 & 85.9 \\
\midrule
\multicolumn{14}{l}{\textbf{Domain transfer (\textbf{ours})}} \\
- \textsc{zshot}\textsuperscript{$\spadesuit$} & 90.5 & 89.6 & 89.9 & 89.2 & 89.3 & 88.4 & 89.8 & 87.5 & 88.7 & 83.6 & 82.8 & 86.7 & 89.2 \\
\phantom{-} \textsc{+trans.} & 91.1 & - & - & - & - & - & - & - & - & - & - & - & - \\
\midrule
\multicolumn{14}{l}{\textbf{Summarization (\textbf{ours})}} \\
- \textsc{zshot}\textsuperscript{$\spadesuit$} & 91.4 & 89.5 & 90.3 & 89.9 & 89.5 & 89.1 & 88.8 & 88.3 & 88.8 & 83.6 & 81.7 & 85.1 & 89.7 \\
\phantom{-} \textsc{+sum.} & 89.9 & - & - & - & - & - & - & - & - & - & - & - & - \\
\bottomrule
\end{tabular}
\end{subtable}
\caption{
\textbf{In-distribution} accuracies for LaBSE, mBERT, and XLM-R. $\spadesuit$: ablations. Scores for \emph{translate-test} are omitted due to the English ID test sets being translated into the respective non-English languages. Note, for English, \textsc{ttrain} does not involve any translation, hence its \en\ scores are equivalent to \textsc{zshot}.}
\label{table:ID_full}
\end{table*}

\begin{table*}
\begin{subtable}{\textwidth}
\scriptsize
\centering
\addtolength{\tabcolsep}{-5pt}
\renewcommand{\arraystretch}{0.88}
\begin{tabular}{l lllllll lllllll lllllllll}
\toprule
\toprule
& \multicolumn{20}{c}{\textbf{LaBSE}} \\ \midrule
& \multicolumn{7}{c}{\textsc{imdb} $\to$ \textsc{amazon}} & \multicolumn{7}{c}{\textsc{imdb} $\to$ \textsc{restaurants}} & \multicolumn{9}{c}{\textsc{imdb} $\to$ \textsc{tweets}} \\
\cmidrule(lr){2-7}  \cmidrule(lr){9-14} \cmidrule(lr){16-23} 
Method & \en & \de & \fr & \es & \ja & \zh & \textbf{\textsc{avg.}} & \en & \dutch & \fr & \es & \ru & \tu & \textbf{\textsc{avg.}} & \en & \de & \fr & \es & \ar & \hi & \italian & \pt & \textbf{\textsc{avg.}} \\ \midrule
\multicolumn{11}{l}{\textbf{Original only}} \\
- \textsc{zshot} & 66.3 & 75.3 & 70.6 & 70.0 & 69.5 & 73.9 & 71.9 & 72.7 & 75.0 & 73.6 & 74.9 & 74.6 & 72.6 & 74.1 & 76.3 & 70.5 & 67.6 & 72.3 & 60.3 & 61.6 & 72.1 & 70.2 & 67.8 \\
- \textsc{ttrain} & 66.3 & 71.6 & 74.2 & 72.5 & 77.0 & 74.8 & 74.0 & 72.7 & 76.2 & 77.5 & 76.7 & 76.1 & 75.4 & 76.4 & 76.3 & 66.3 & 67.1 & 70.1 & 56.1 & 62.3 & 69.3 & 71.1 & 66.0 \\
- \textsc{ttest} & 66.3 & 70.0 & 67.6 & 66.4 & 66.4 & 67.5 & 67.6 & 72.7 & 75.6 & 72.5 & 73.8 & 70.4 & 73.3 & 73.1 & 76.3 & 70.6 & 64.8 & 72.4 & 60.6 & 67.7 & 73.3 & 72.4 & 68.8 \\
\midrule
\multicolumn{11}{l}{\textbf{Original \& CAD} \cite{kaushik2019learning}} \\
- \textsc{zshot} & 81.2 & 85.4 & 85.3 & 85.0 & 80.4 & 78.5 & \underline{82.9} & 84.7 & 86.8 & 86.4 & 88.6 & 83.5 & 83.3 & 85.7 & 81.7 & 76.6 & 72.2 & 80.3 & 71.6 & 67.8 & 75.2 & 77.8 & \underline{74.5} \\
- \textsc{ttrain} & 81.2  & 85.0 & 83.5 & 84.5 & 80.0 & 78.7 & 82.3 & 84.7 & 84.4 & 81.6 & 88.6 & 80.8 & 81.5 & 83.4 & 81.7 & 77.6 & 72.6 & 81.0 & 67.4 & 64.7 & 74.8 & 77.8 & 73.7 \\
- \textsc{ttest} & 81.2 & 84.4 & 84.9 & 83.7 & 79.8 & 79.3 & 82.4 & 84.7 & 88.0 & 86.4 & 87.9 & 82.2 & 85.0 & 85.9 & \textbf{81.7} & 79.8 & 71.7 & 81.6 & 71.0 & 74.8 & 75.0 & 79.3 & \textbf{76.2} \\
\midrule
\multicolumn{11}{l}{\textbf{Original \& CORE} \cite{dixit-etal-2022-core}} \\
- \textsc{zshot} & 81.0 & 84.8 & 84.2 & 84.6 & 80.2 & 76.3 & 82.0 & 85.0 & 84.6 & 85.4 & 88.7 & 84.7 & 81.2 & 84.9 & 77.4 & 71.2 & 67.6 & 76.0 & 66.9 & 64.0 & 75.7 & 76.0 & 71.1 \\
- \textsc{ttest} & \cellcolor{green!10}81.0 & \cellcolor{green!10}84.4 & \cellcolor{green!10}83.9 & \cellcolor{green!10}83.2 & \cellcolor{green!10}79.8 & \cellcolor{green!10}77.1 & \cellcolor{green!10}81.7 & \cellcolor{green!10}\underline{85.0} & \cellcolor{green!10}86.5 & \cellcolor{green!10}85.3 & \cellcolor{green!10}89.5 & \cellcolor{green!10}84.1 & \cellcolor{green!10}86.2 & \cellcolor{green!10}\underline{86.3} & \cellcolor{green!10}\underline{77.4} & \cellcolor{green!10}77.9 & \cellcolor{green!10}69.8 & \cellcolor{green!10}80.5 & \cellcolor{green!10}65.3 & \cellcolor{green!10}72.8 & \cellcolor{green!10}76.0 & \cellcolor{green!10}77.8 & \cellcolor{green!10}74.3 \\
\midrule
\multicolumn{11}{l}{\textbf{Original \& Domain transfer (\textbf{ours})}} \\
- \textsc{zshot}\textsuperscript{$\spadesuit$} & 76.0 & 82.5 & 79.5 & 79.1 & 77.7 & 75.8 & 78.9 & 81.4 & 83.1 & 81.2 & 82.6 & 82.0 & 78.4 & 81.5 & 80.9 & 72.3 & 68.1 & 76.2 & 65.2 & 64.7 & 74.8 & 74.3 & 70.8 \\
\phantom{-} \textsc{+ttest}\textsuperscript{$\spadesuit$} & 76.0 & 80.6 & 79.8 & 79.2 & 76.6 & 75.6 & 78.4 & 81.4 & 84.4 & 82.8 & 81.6 & 80.7 & 81.6 & 82.2 & 80.9 & 72.7 & 69.1 & 75.4 & 66.3 & 74.1 & 74.3 & 74.3 & 72.3 \\
\phantom{-} \textsc{+tran.} &
\underline{81.7} & 83.6 & 83.7 & 83.0 & 81.1 & 78.0 & 81.9 & 84.1 & 85.9 & 84.2 & 85.2 & 83.1 & 82.1 & 84.1 & 72.3 & 69.1 & 62.0 & 74.9 & 62.6 & 71.0 & 71.1 & 76.5 & 69.6 \\
\midrule
\multicolumn{11}{l}{\textbf{Original \& Summarization (\textbf{ours})}} \\
- \textsc{zshot}\textsuperscript{$\spadesuit$} & 77.1 & 82.5 & 80.7 & 81.2 & 77.8 & 76.2 & 79.7 & 83.6 & 85.2 & 83.7 & 84.7 & 84.2 & 80.5 & 83.7 & 81.9 & 73.4 & 70.9 & 77.9 & 65.0 & 66.2 & 75.5 & 74.0 & 71.8 \\
\phantom{-} \textsc{+ttest}\textsuperscript{$\spadesuit$} & 77.1 & 81.1 & 80.4 & 80.2 & 76.6 & 76.1 & 78.9 & 83.6 & 86.7 & 83.5 & 83.0 & 84.1 & 82.6 & 84.0 & 81.9 & 74.7 & 69.3 & 77.6 & 68.1 & 75.3 & 73.1 & 73.4 & 73.1 \\
\phantom{-} \textsc{+sum.} & \cellcolor{green!10}\textbf{86.2}\win & \cellcolor{green!10}86.3\win & \cellcolor{green!10}87.6\win & \cellcolor{green!10}87.5\win & \cellcolor{green!10}82.6\win & \cellcolor{green!10}79.7\win & \cellcolor{green!10}\textbf{84.7} & \cellcolor{green!10}\textbf{91.6}\win & \cellcolor{green!10}89.5\win & \cellcolor{green!10}89.1\win & \cellcolor{green!10}89.5\phantom{\win} & \cellcolor{green!10}89.2\win & \cellcolor{green!10}86.5\win & \cellcolor{green!10}\textbf{88.8} & \cellcolor{green!10}76.6\lose & \cellcolor{green!10}74.7\lose & \cellcolor{green!10}73.3\win & \cellcolor{green!10}81.0\win & \cellcolor{green!10}70.2\win & \cellcolor{green!10}74.3\win & \cellcolor{green!10}71.7\lose & \cellcolor{green!10}73.1\lose & \cellcolor{green!10}74.0 \\
\midrule
\end{tabular}
\end{subtable}
\begin{subtable}{\textwidth}
\scriptsize
\centering
\addtolength{\tabcolsep}{-5pt}
\renewcommand{\arraystretch}{0.88}
\begin{tabular}{l lllllll lllllll lllllllll}
\toprule
\toprule
& \multicolumn{20}{c}{\textbf{mBERT}} \\ \midrule
& \multicolumn{7}{c}{\textsc{imdb} $\to$ \textsc{amazon}} & \multicolumn{7}{c}{\textsc{imdb} $\to$ \textsc{restaurants}} & \multicolumn{9}{c}{\textsc{imdb} $\to$ \textsc{tweets}} \\
\cmidrule(lr){2-7}  \cmidrule(lr){9-14} \cmidrule(lr){16-23} 
Method & \en & \de & \fr & \es & \ja & \zh & \textbf{\textsc{avg.}} & \en & \dutch & \fr & \es & \ru & \tu & \textbf{\textsc{avg.}} & \en & \de & \fr & \es & \ar & \hi & \italian & \pt & \textbf{\textsc{avg.}} \\ \midrule
\multicolumn{11}{l}{\textbf{Original only}} \\
- \textsc{zshot} & 79.3 & 72.2 & 73.1 & 74.5 & 71.6 & 69.8 & 72.2 & 80.2 & 69.8 & 68.8 & 72.2 & 73.3 & 64.1 & 69.6 & 75.9 & 60.5 & 66.2 & 64.0 & 61.4 & 58.3 & 65.8 & 63.4 & 62.8 \\
- \textsc{ttrain} & 79.3 & 72.6 & 77.6 & 76.8 & 71.0 & 69.4 & 73.5 & 80.2 & 75.4 & 75.3 & 78.4 & 76.8 & 66.7 & 74.5 & 75.9 & 57.7 & 69.5 & 66.7 & 64.3 & 52.6 & 66.9 & 62.4 & 62.9 \\
- \textsc{ttest} & 79.3 & 78.9 & 79.8 & 80.3 & 75.2 & 74.6 & 77.8 & 80.2 & 79.4 & 78.2 & 82.2 & 79.2 & 75.4 & 78.9 & 75.9 & 67.4 & 67.1 & 73.8 & 68.3 & 72.0 & 73.5 & 75.7 & 71.1 \\
\midrule
\multicolumn{11}{l}{\textbf{Original \& CAD} \cite{kaushik2019learning}} \\
- \textsc{zshot} & 81.7 & 76.0 & 76.0 & 77.7 & 73.1 & 71.9 & 74.9 & 81.8 & 68.6 & 71.2 & 77.1 & 72.7 & 64.9 & 70.9 & 79.0 & 64.3 & 74.9 & 68.9 & 69.0 & 61.0 & 68.3 & 64.2 & 67.2 \\
- \textsc{ttrain} & 81.7 & 79.0 & 80.5 & 80.4 & 76.5 & 74.5 & 78.2 & 81.8 & 75.9 & 76.6 & 81.5 & 74.5 & 69.9 & 75.7 & 79.0 & 64.9 & 75.6 & 71.2 & 65.0 & 54.8 & 70.4 & 66.7 & 66.9 \\
- \textsc{ttest} & \textbf{81.7} & 82.7 & 83.3 & 83.2 & 79.4 & 77.4 & \textbf{81.2} & 81.8 & 81.4 & 81.5 & 83.9 & 79.1 & 79.9 & \underline{81.2} & \textbf{79.0} & 73.9 & 74.1 & 78.3 & 75.5 & 73.3 & 72.6 & 77.5 & \textbf{75.0} \\
\midrule
\multicolumn{11}{l}{\textbf{Original \& CORE} \cite{dixit-etal-2022-core}} \\
- \textsc{zshot} & 80.2 & 74.3 & 75.3 & 77.2 & 73.6 & 70.2 & 74.1 & 80.4 & 65.3 & 72.1 & 75.3 & 71.2 & 63.9 & 69.6 & 73.6 & 59.4 & 72.0 & 70.3 & 62.7 & 59.3 & 68.3 & 61.5 & 64.8 \\
- \textsc{ttest} & \cellcolor{green!10}80.7 & \cellcolor{green!10}81.3 & \cellcolor{green!10}80.4 & \cellcolor{green!10}82.5 & \cellcolor{green!10}79.2 & \cellcolor{green!10}76.3 & \cellcolor{green!10}79.9 & \cellcolor{green!10}80.4 & \cellcolor{green!10}79.2 & \cellcolor{green!10}79.7 & \cellcolor{green!10}82.9 & \cellcolor{green!10}78.5 & \cellcolor{green!10}79.4 & \cellcolor{green!10}79.9 & \cellcolor{green!10}73.6 & \cellcolor{green!10}70.6 & \cellcolor{green!10}70.0 & \cellcolor{green!10}77.9 & \cellcolor{green!10}73.0 & \cellcolor{green!10}70.1 & \cellcolor{green!10}73.0 & \cellcolor{green!10}75.1 & \cellcolor{green!10}72.8 \\
\midrule
\multicolumn{11}{l}{\textbf{Original \& Domain transfer (\textbf{ours})}} \\
- \textsc{zshot}\textsuperscript{$\spadesuit$} & 79.6 & 73.2 & 74.8 & 76.4 & 72.3 & 71.0 & 73.5 & 80.2 & 70.8 & 70.4 & 73.6 & 73.1 & 63.9 & 70.4 & 78.1 & 60.5 & 69.0 & 63.8 & 62.6 & 58.8 & 66.0 & 64.9 & 63.7 \\
\phantom{-} \textsc{+ttest}\textsuperscript{$\spadesuit$} & 79.6 & 80.3 & 81.0 & 80.8 & 76.8 & 75.8 & 78.9 & 80.2 & 78.2 & 77.8 & 80.9 & 78.3 & 76.4 & 78.3 & \underline{78.1} & 68.8 & 68.2 & 73.9 & 72.3 & 72.7 & 72.9 & 75.6 & 72.1 \\
\phantom{-} \textsc{+tran.} & 81.3 & 81.4 & 81.6 & 81.9 & 79.5 & 77.0 & \underline{80.3} & \underline{83.3} & 81.0 & 80.4 & 83.6 & 80.4 & 79.6 & 81.0 & 72.4 & 67.5 & 66.2 & 72.2 & 65.1 & 70.6 & 70.3 & 75.9 & 69.7 \\
\midrule
\multicolumn{11}{l}{\textbf{Original \& Summarization (\textbf{ours})}} \\
- \textsc{zshot}\textsuperscript{$\spadesuit$} & 80.7 & 74.1 & 75.4 & 77.1 & 72.3 & 69.2 & 73.6 & 82.4 & 71.1 & 72.4 & 76.8 & 75.5 & 66.6 & 72.5 & 77.8 & 60.6 & 67.1 & 66.8 & 61.5 & 59.5 & 65.3 & 63.8 & 63.5 \\
\phantom{-} \textsc{+ttest}\textsuperscript{$\spadesuit$} & 80.7 & 81.5 & 82.3 & 82.4 & 76.7 & 75.3 & 79.6 & 82.4 & 80.0 & 80.2 & 83.0 & 79.7 & 77.3 & 80.0 & 77.8 & 70.1 & 67.5 & 75.6 & 70.8 & 72.4 & 71.4 & 76.2 & 72.0 \\
\phantom{-} \textsc{+sum.} & \cellcolor{green!10}\underline{81.0}\win & \cellcolor{green!10}82.3\win & \cellcolor{green!10}83.6\win & \cellcolor{green!10}84.0\win & \cellcolor{green!10}78.1\lose & \cellcolor{green!10}77.8\win & \cellcolor{green!10}\textbf{81.2} & \cellcolor{green!10}\textbf{87.3}\win & \cellcolor{green!10}84.6\win & \cellcolor{green!10}85.5\win & \cellcolor{green!10}87.3\win & \cellcolor{green!10}83.6\win & \cellcolor{green!10}80.4\win & \cellcolor{green!10}\textbf{84.3} & \cellcolor{green!10}74.3\win & \cellcolor{green!10}73.0\win & \cellcolor{green!10}72.1\win & \cellcolor{green!10}76.9\lose & \cellcolor{green!10}76.1\win & \cellcolor{green!10}71.6\win & \cellcolor{green!10}69.9\lose & \cellcolor{green!10}77.0\win & \cellcolor{green!10}\underline{73.8} \\
\midrule
\end{tabular}
\end{subtable}
\begin{subtable}{\textwidth}
\scriptsize
\centering
\addtolength{\tabcolsep}{-5pt}
\renewcommand{\arraystretch}{0.88}
\begin{tabular}{l lllllll lllllll lllllllll}
\toprule
\toprule
& \multicolumn{20}{c}{\textbf{XLM-R}} \\ \midrule
& \multicolumn{7}{c}{\textsc{imdb} $\to$ \textsc{amazon}} & \multicolumn{7}{c}{\textsc{imdb} $\to$ \textsc{restaurants}} & \multicolumn{9}{c}{\textsc{imdb} $\to$ \textsc{tweets}} \\
\cmidrule(lr){2-7}  \cmidrule(lr){9-14} \cmidrule(lr){16-23} 
Method & \en & \de & \fr & \es & \ja & \zh & \textbf{\textsc{avg.}} & \en & \dutch & \fr & \es & \ru & \tu & \textbf{\textsc{avg.}} & \en & \de & \fr & \es & \ar & \hi & \italian & \pt & \textbf{\textsc{avg.}} \\ \midrule
\multicolumn{11}{l}{\textbf{Original only}} \\
- \textsc{zshot} & 86.3 & 86.7 & 85.0 & 83.9 & 86.9 & 82.4 & 85.0 & 86.0 & 81.2 & 78.6 & 80.7 & 81.9 & 73.4 & 79.2 & 84.3 & 75.5 & 66.0 & 72.9 & 68.4 & 63.6 & 70.0 & 68.0 & 69.2 \\
- \textsc{ttrain} & 86.3 & 86.9 & 86.5 & 88.2 & 87.1 & 81.4 & 86.0 & 86.0 & 85.9 & 79.2 & 86.7 & 85.5 & 77.9 & 83.0 & 84.3 & 75.4 & 66.9 & 82.1 & 71.3 & 66.6 & 71.6 & 73.6 & 72.5 \\
- \textsc{ttest} & 86.3 & 86.7 & 87.8 & 86.6 & 85.5 & 81.4 & 85.6 & 86.0 & 81.6 & 82.2 & 86.0 & 79.8 & 79.8 & 81.5 & 84.3 & 76.6 & 67.5 & 77.3 & 70.2 & 70.0 & 69.4 & 71.2 & 71.7 \\
\midrule
\multicolumn{11}{l}{\textbf{Original \& CAD} \cite{kaushik2019learning}} \\
- \textsc{zshot} & 87.0 & 86.9 & 86.3 & 86.3 & 86.2 & 82.7 & 85.7 & 87.5 & 82.5 & 81.8 & 83.3 & 82.1 & 79.6 & 81.9 & 86.7 & 77.6 & 76.1 & 82.7 & 78.2 & 67.9 & 74.2 & 74.6 & 75.9 \\
- \textsc{ttrain} & 87.0 & 87.6 & 87.8 & 88.4 & 87.0 & 81.0 & 86.4 & 87.5 & 85.3 & 83.5 & 87.6 & 85.0 & 81.7 & 84.6 & 86.7 & 80.4 & 75.1 & 85.0 & 79.6 & 68.4 & 75.6 & 77.0 & 77.3 \\
- \textsc{ttest} & 87.0 & 87.8 & 88.8 & 88.4 & 86.9 & 82.1 & 86.8 & 87.5 & 87.3 & 86.5 & 89.2 & 85.8 & 86.9 & 87.1 & \textbf{86.7} & 81.4 & 77.6 & 84.3 & 79.6 & 76.0 & 77.8 & 80.6 & \underline{79.6} \\
\midrule
\multicolumn{11}{l}{\textbf{Original \& CORE} \cite{dixit-etal-2022-core}} \\
- \textsc{zshot} & 86.8 & 88.1 & 87.7 & 88.7 & 88.9 & 81.6 & 87.0 & 89.7 & 88.8 & 87.2 & 90.4 & 89.1 & 81.9 & 87.5 & 83.9 & 75.7 & 79.4 & 82.9 & 80.9 & 67.8 & 79.9 & 78.8 & 77.9 \\
- \textsc{ttest} & \cellcolor{green!10}86.8 & \cellcolor{green!10}88.4 & \cellcolor{green!10}89.0 & \cellcolor{green!10}89.0 & \cellcolor{green!10}87.6 & \cellcolor{green!10}81.1 & \cellcolor{green!10}87.0 & \cellcolor{green!10}\underline{89.7} & \cellcolor{green!10}89.2 & \cellcolor{green!10}89.0 & \cellcolor{green!10}91.2 & \cellcolor{green!10}88.0 & \cellcolor{green!10}88.1 & \cellcolor{green!10}\underline{89.1} & \cellcolor{green!10}83.9 & \cellcolor{green!10}81.1 & \cellcolor{green!10}77.6 & \cellcolor{green!10}86.2 & \cellcolor{green!10}82.2 & \cellcolor{green!10}75.4 & \cellcolor{green!10}79.6 & \cellcolor{green!10}81.2 & \cellcolor{green!10}\textbf{80.5} \\
\midrule
\multicolumn{11}{l}{\textbf{Original \& Domain transfer (\textbf{ours})}} \\
- \textsc{zshot}\textsuperscript{$\spadesuit$} & 86.4 & 86.9 & 85.5 & 84.6 & 87.1 & 82.0 & 85.2 & 85.4 & 80.1 & 79.2 & 81.7 & 82.3 & 74.4 & 79.5 & 85.2 & 75.7 & 69.2 & 75.6 & 70.6 & 65.6 & 71.1 & 69.7 & 71.1 \\
\phantom{-} \textsc{+ttest}\textsuperscript{$\spadesuit$} & 86.4 & 88.1 & 89.0 & 88.0 & 87.5 & 81.7 & 86.9 & 85.4 & 84.0 & 83.4 & 85.7 & 83.0 & 83.7 & 84.0 & 85.2 & 78.4 & 71.8 & 80.4 & 74.9 & 73.8 & 73.5 & 74.7 & 75.4 \\
\phantom{-} \textsc{+tran.} & \underline{87.1} & 88.3 & 89.2 & 88.4 & 87.1 & 82.5 & \underline{87.1} & 87.2 & 84.3 & 85.0 & 87.0 & 82.8 & 83.4 & 84.5 & 72.7 & 72.4 & 66.0 & 73.7 & 65.8 & 70.0 & 66.4 & 73.9 & 69.7 \\
\midrule
\multicolumn{11}{l}{\textbf{Original \& Summarization (\textbf{ours})}} \\
- \textsc{zshot}\textsuperscript{$\spadesuit$} & 87.8 & 89.1 & 89.3 & 88.7 & 88.1 & 83.3 & 87.7 & 89.4 & 86.1 & 83.8 & 86.5 & 86.5 & 81.7 & 84.9 & 86.3 & 76.6 & 71.7 & 81.6 & 75.8 & 69.0 & 75.7 & 75.2 & 75.1 \\
\phantom{-} \textsc{+ttest}\textsuperscript{$\spadesuit$} & 87.8 & 89.5 & 90.5 & 89.5 & 88.0 & 82.4 & \textbf{88.0} & 89.4 & 87.5 & 87.7 & 88.6 & 85.8 & 85.7 & 87.1 & \underline{86.3} & 79.8 & 73.7 & 83.0 & 77.1 & 75.7 & 75.1 & 80.4 & 77.8 \\
\phantom{-} \textsc{+sum.} & \cellcolor{green!10}\textbf{87.8}\win & \cellcolor{green!10}87.6\lose & \cellcolor{green!10}89.7\win & \cellcolor{green!10}89.2\win & \cellcolor{green!10}86.1\lose & \cellcolor{green!10}81.2\win & \cellcolor{green!10}86.8\lose & \cellcolor{green!10}\textbf{92.8}\win & \cellcolor{green!10}91.0\win & \cellcolor{green!10}90.1\win & \cellcolor{green!10}91.8\win & \cellcolor{green!10}89.5\win & \cellcolor{green!10}88.8\win & \cellcolor{green!10}\textbf{90.2} & \cellcolor{green!10}83.0\lose & \cellcolor{green!10}78.0\lose & \cellcolor{green!10}74.6\lose & \cellcolor{green!10}80.0\lose & \cellcolor{green!10}76.0\lose & \cellcolor{green!10}74.1\lose & \cellcolor{green!10}71.4\lose & \cellcolor{green!10}77.0\lose & \cellcolor{green!10}75.9 \\
\midrule
\end{tabular}
\end{subtable}
\caption{
\textbf{Out-of-distribution} accuracies for LaBSE, mBERT, and XLM-R.  \textbf{Best} model 
in bold with the \underline{runner-up} underlined. $\spadesuit$: ablations. For English, \textsc{ttrain} and \textsc{ttest} do not involve any translation, hence their \en\ scores are equivalent to \textsc{zshot}. \colorbox{green!10}{Highlighted} rows show a 1-on-1 comparison between classifiers augmented with
\begin{enumerate*}[(i)]
    \item our (\emph{summarization}) strategy, and
    \item the state-of-the-art generated CORE counterfactuals. 
\end{enumerate*}}
\label{table:OOD_full}
\end{table*}

\end{document}